\documentclass[sigconf]{acmart}
\usepackage{popets}
\usepackage{soul}
\usepackage{makecell}
\setcopyright{popets}
\copyrightyear{YYYY}

\acmYear{2025}
\acmVolume{2025}
\acmNumber{X}
\acmDOI{XXXXXXX.XXXXXXX}
\acmISBN{}
\acmConference{Proceedings on Privacy Enhancing Technologies}
\usepackage{graphicx} 
\usepackage{subcaption} 
\usepackage{multirow}
\usepackage{hyperref}
\hypersetup{
    colorlinks=true,
    linkcolor=blue,
    filecolor=magenta,      
    urlcolor=cyan,
    pdftitle={Overleaf Example},
    pdfpagemode=FullScreen,
    }

\begin{document}\sloppy

\title[The Interplay Among Privacy, Utility, and Fairness in Image Classification]{The Impact of Generalization Techniques on the Interplay Among Privacy, Utility, and Fairness in Image Classification}


\author{Ahmad Hassanpour}
\orcid{1234-5678-9012}
\affiliation{%
  \institution{Norwegian University of Science and Technology}
  \city{}
  \state{}
  \country{}}
\email{ahmad.hassanpour@ntnu.no}

\author{Amir Zarei}
\affiliation{%
  \institution{Norwegian University of Science and Technology}
  \city{}
  \country{}}
\email{amir.zarei@ntnu.no}

\author{Khawla Mallat}
\affiliation{%
  \institution{SAP Labs France}
  \city{}
  \country{}
}
\email{khawla.mallat@sap.com}

\author{Anderson Santana de Oliveira}
\affiliation{%
 \institution{SAP Labs France}
 \city{}
 \state{}
 \country{}}
\email{anderson.santana.de.oliveira@sap.com}

\author{Bian Yang}
\affiliation{%
  \institution{Norwegian University of Science and Technology}
  \city{}
  \state{}
  \country{}}
\email{bian.yang@ntnu.no}


\renewcommand{\shortauthors}{Hassanpour et al.}

\begin{abstract}

This study investigates the trade-offs between fairness, privacy, and utility in image classification using machine learning (ML). Recent research suggests that generalization techniques can improve the balance between privacy and utility. One focus of this work is sharpness-aware training (SAT) and its integration with differential privacy (DP-SAT) to further improve this balance. Additionally, we examine fairness in both private and non-private learning models trained on datasets with synthetic and real-world biases. We also measure the privacy risks involved in these scenarios by performing membership inference attacks (MIAs) and explore the consequences of eliminating high-privacy risk samples, termed outliers. Moreover, we introduce a new metric, named \emph{harmonic score}, which combines accuracy, privacy, and fairness into a single measure.

Through empirical analysis using generalization techniques, we achieve an accuracy of 81.11\% under $(8, 10^{-5})$-DP on CIFAR-10, surpassing the 79.5\% reported by De et al. (2022). Moreover, our experiments show that memorization of training samples can begin before the overfitting point, and generalization techniques do not guarantee the prevention of this memorization. Our analysis of synthetic biases shows that generalization techniques can amplify model bias in both private and non-private models. Additionally, our results indicate that increased bias in training data leads to reduced accuracy, greater vulnerability to privacy attacks, and higher model bias. We validate these findings with the CelebA dataset, demonstrating that similar trends persist with real-world attribute imbalances. Finally, our experiments show that removing outlier data decreases accuracy and further amplifies model bias.
\end{abstract}


\maketitle
\section{Introduction}
\label{sec:intro}
Privacy and fairness are important elements in developing responsible machine learning (ML) models. Privacy ensures that individual data contributions remain confidential and are not identifiable in the model’s outputs. On the other hand, fairness involves ensuring that the model’s outputs are unbiased and equitable across various demographic groups, preventing discrimination and ensuring inclusivity. While significant progress has been made in understanding and addressing individual trade-offs, such as the balance between privacy and utility and the balance between fairness and utility, the interplay between these trade-offs has not been thoroughly investigated. This gap is particularly evident in image classification, where models need to handle complex and diverse data inputs. Understanding how privacy and fairness affect each other in this domain is essential for creating ML models that are both secure and equitable.

Differential privacy (DP) is a gold standard for ensuring privacy in ML models, offering mathematical guarantees that individual data entries in aggregated datasets remain protected. Introduced by Dwork et al.~\cite{dwork2006calibrating}, DP works by adding controlled noise to the data, which masks the contributions of individual entries. This noise makes identifying any single data point within the dataset difficult, thus safeguarding personal information. Despite this protection, DP still preserves the overall patterns in the data, allowing ML models to be trained effectively. However, this approach also highlights a significant challenge: the privacy-utility trade-off. Adding more noise increases privacy but can also reduce the accuracy of the ML model, thereby lowering its utility. Thus, balancing privacy and utility is challenging for successfully implementing DP in ML models.

A set of generalization techniques, including group normalization, optimal batch size, weight standardization, augmentation multiplicity, and parameter averaging, have been shown to significantly enhance the utility of deep ML models trained using differentially private stochastic gradient descent (DP-SGD)~\cite{de2022unlocking}. However, the impact of a recently proposed generalization technique, differentially private sharpness-aware training (DP-SAT)~\cite{park2023differentially}, which can serve as an alternative to DP-SGD, has not been thoroughly investigated. This raises an important question: \textbf{Q1: Would combining DP-SAT with the other generalization techniques lead to an even better utility-privacy trade-off?}

Fairness in ML models can be defined as the goal of producing unbiased and equitable predictions across demographic groups. However, model bias may occur when systematic errors or prejudices arise in predictions, potentially disadvantaging certain groups. Research shows that when training data is biased, models trained on it tend to learn and incorporate these biases into their predictions~\cite{10.1145/3457607}. While DP seeks to safeguard individual privacy, its application can inadvertently compromise fairness in deep ML models~\cite{bagdasaryan2019differential,ganev2022robin,chang2021privacy,uniyal2021dp, de2023empirical}. Recent studies~\cite{bagdasaryan2019differential,ganev2022robin} indicate that the noise introduced by DP measures can disproportionately impact underrepresented groups within a dataset. This leads to a greater decrease in accuracy for these groups compared to their well-represented counterparts, thereby generating biased outcomes. However, the impact of the generalization techniques on fairness is an unexplored area to our knowledge. These techniques could potentially exacerbate model bias in scenarios with and without DP. \textbf{Q2: Would the generalization techniques impact ML model fairness in non-private and private settings?}

Evaluating ML models has evolved beyond simply assessing their utility to include other aspects, such as privacy and fairness. This needs a multidimensional evaluation framework that integrates three important aspects: privacy, utility, and fairness. This holistic evaluation approach requires that models are not only efficient in their intended tasks but also comply with the principles of privacy protection and ensuring equitable outcomes across diverse demographic groups. Thus, this shift requires that evaluation metrics be capable of simultaneously capturing the nuances of utility, privacy, and fairness. \textbf{Q3: How could we integrate utility, privacy, and fairness in a single metric?}


In this paper, we address the above-mentioned research questions exploring the impact of generalization techniques on the three-dimensional aspects (i.e., utility, privacy, fairness). Moreover, we measure privacy risks involved by applying various membership inference attacks (MIAs)~\cite{shokri2017Membership, song2021systematic} on ML models trained on unbiased data and data with synthetic and real-world biases. To explore further, we evaluate the impact of removing outlier samples that are most susceptible to privacy attacks on utility, privacy, and fairness in both private and non-private settings.


More specifically, our empirical analysis leads to the following contributions:
\begin{figure*}
    \centering
    \begin{subfigure}{0.3\linewidth}
        \includegraphics[width=\textwidth]{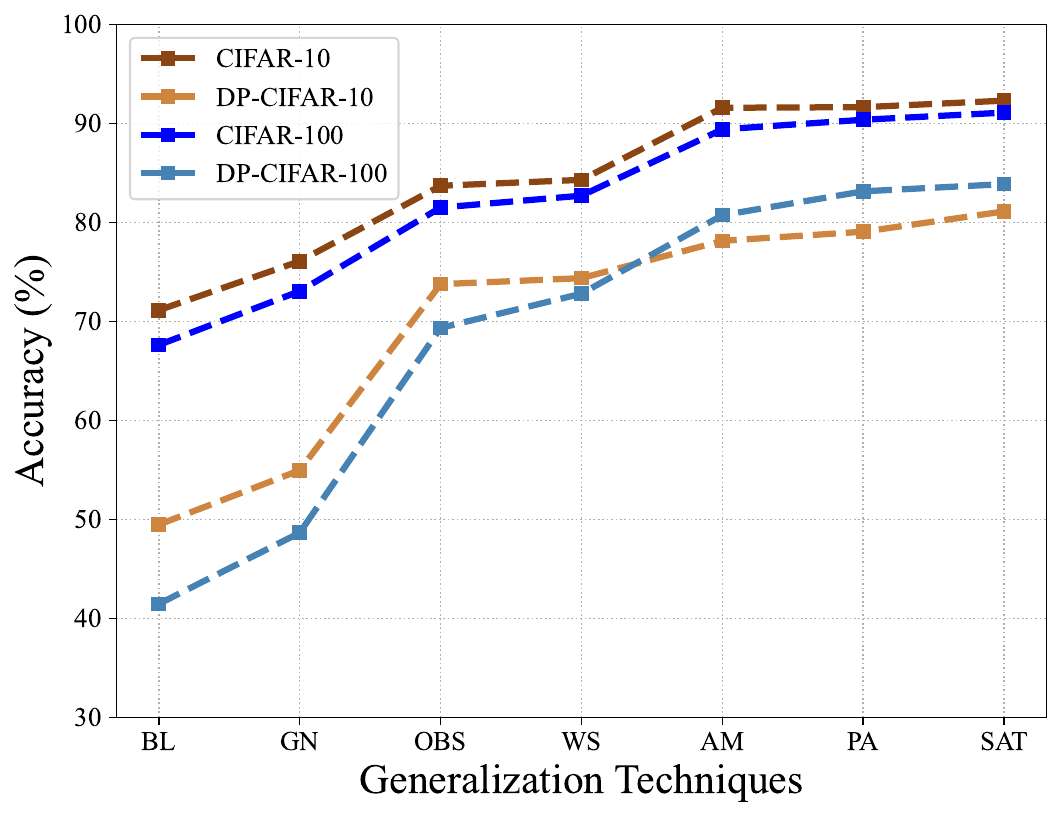}
        \caption{}
        \label{fig:gt_accuracy}
    \end{subfigure}
    \hfill
   \begin{subfigure}{0.3\linewidth}
       \includegraphics[width=\textwidth]{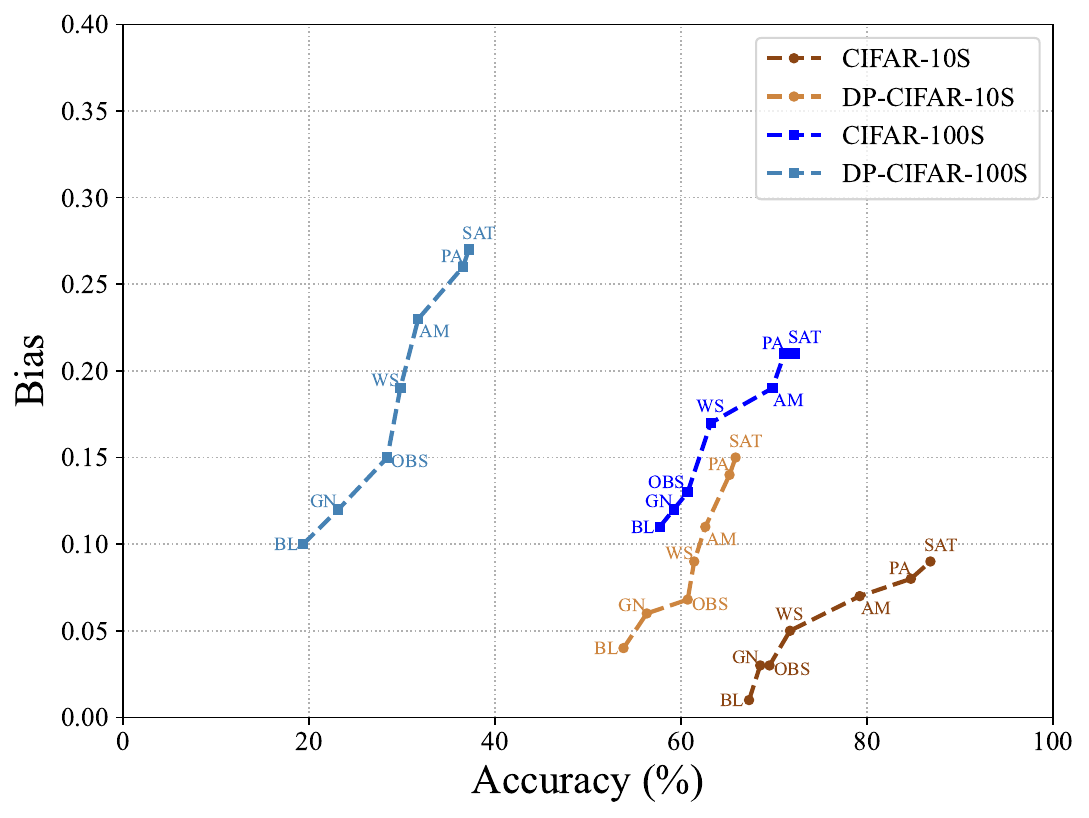}
       \caption{}
        \label{fig:bias_accuracy}
   \end{subfigure}
   \hfill
    \begin{subfigure}{0.3\linewidth}
        \includegraphics[width=\textwidth]{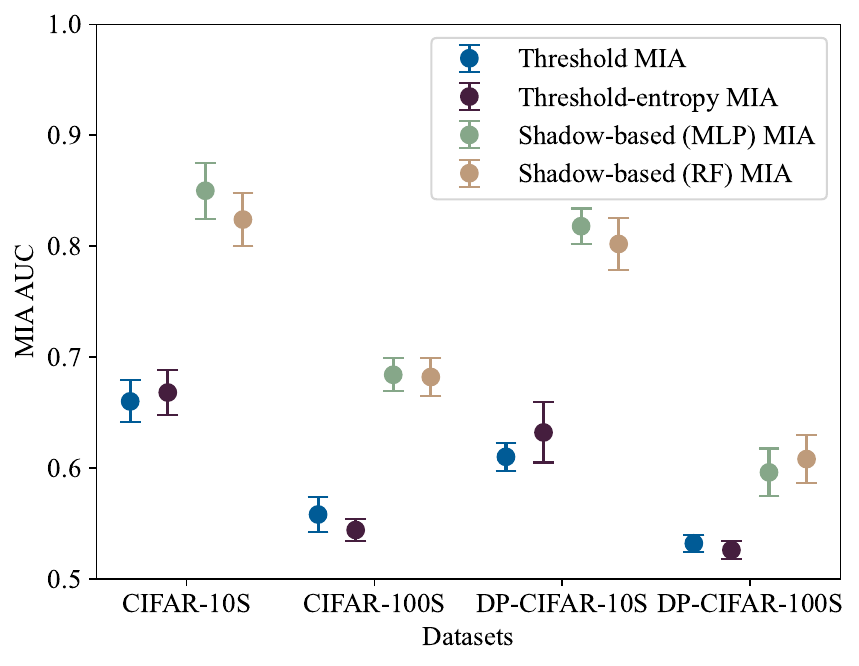}
        \caption{}
        \label{fig:all_mem_acc_cifarS}
    \end{subfigure}
    \hfill
   \begin{subfigure}{0.3\linewidth}
       \includegraphics[width=\textwidth]{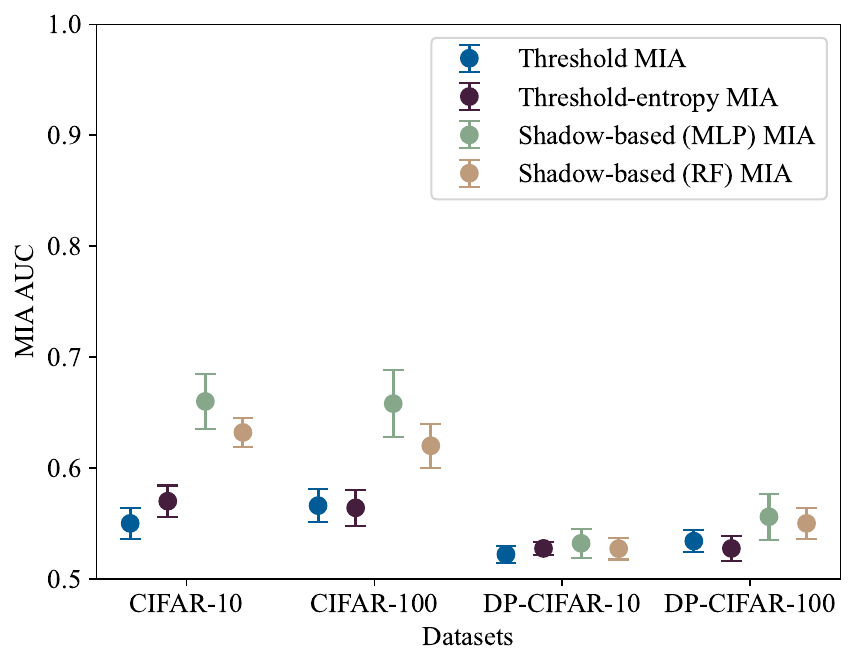}
       \caption{}
        \label{fig:all_mem_acc_cifar}
   \end{subfigure}
   \hfill
   \begin{subfigure}{0.5\linewidth} 
    \centering
    \includegraphics[width=\textwidth,height=0.65\textheight,keepaspectratio]{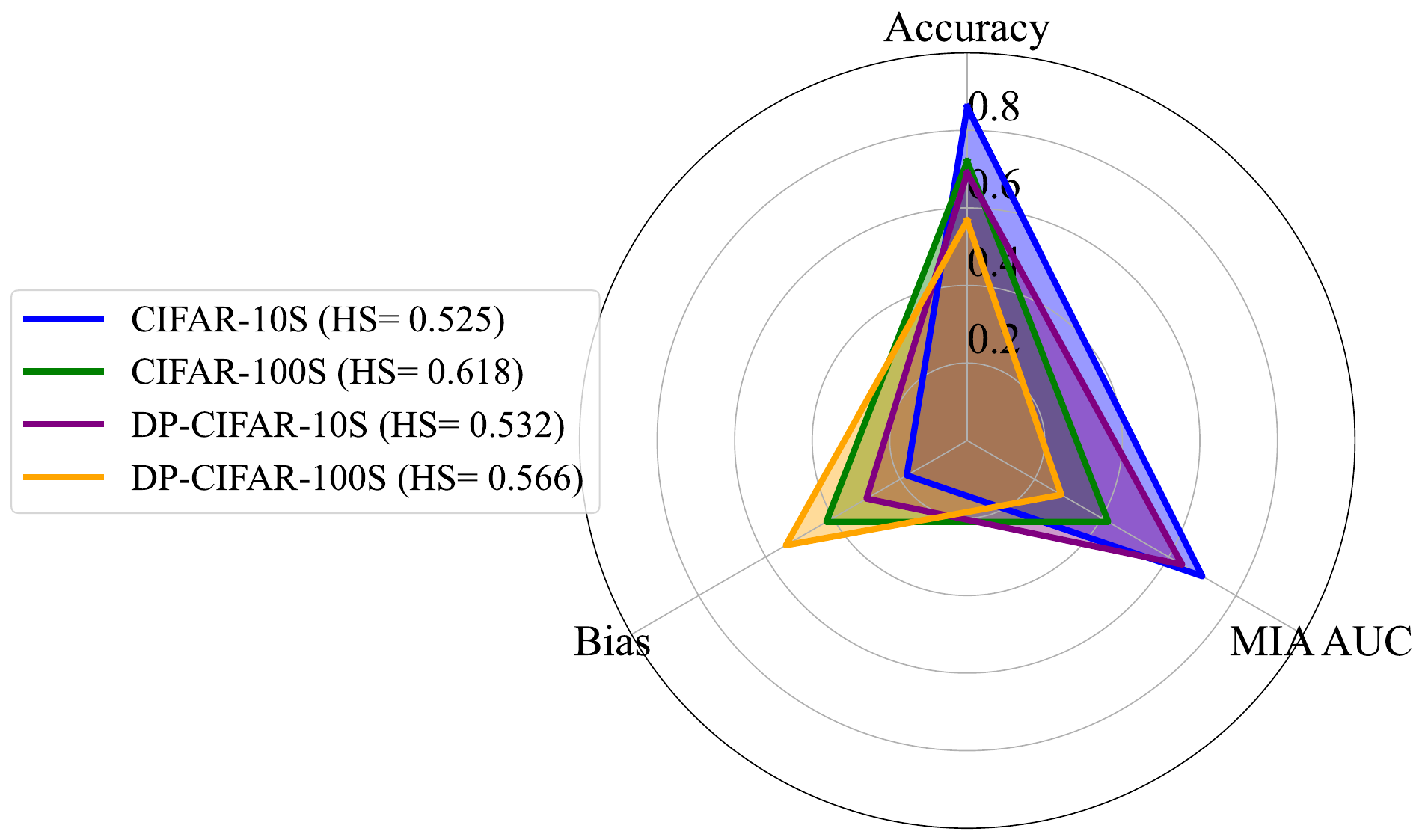}
    \caption{}
    \label{fig:abe}
\end{subfigure}  
    \caption{Variations in accuracy, MIA AUC, and model bias for four datasets CIFAR-10, CIFAR-100, CIFAR-10S, and CIFAR-100S in non-private and private (i.e., $(8, 10^{-5})$-DP) learning settings. DP-CIFAR-10/10S/100/100S is used to denote when DP is applied. \textbf{(a)} illustrates the impact of generalization techniques (i.e., BL: baseline, OBS: optimal batch size, GN: group normalization, WS: weight standardization, AM: augmentation multiplicity, PA: parameter averaging, SAT: sharpness-aware training) on accuracy when the training data is unbiased while \textbf{(b)} measures such an impact on model accuracy and bias when training data is biased. \textbf{(c)} and \textbf{(d)} show MIA AUC of two threshold-based MIAs (i.e., threshold, threshold entropy) and two shadow-based MIAs (MLP: multilayer perceptron, RF: random forest). \textbf{(e)} compares the HS to represent the balance between accuracy, MIA AUC, and bias.}
    \label{fig:overall_f1}
\end{figure*}

\begin{itemize}
\item The generalization techniques suggested by De et al.~\cite{de2022unlocking} have enhanced the balance between accuracy and privacy. We incorporate the DP-SAT method, which further improves this balance. Notably, substituting DP-SGD with DP-SAT resulted in achieving a new accuracy of 81.11\% under $(8, 10^{-5})$-DP using a 16-layer Wide-ResNet without extra data on CIFAR-10, improving the previously reported 79.5\% by~\cite{de2022unlocking} (the accuracy values are reported over the official test set of CIFAR-10). We also show the superior performance of DP-SAT compared to DP-SGD when combined with the generalization techniques for different privacy parameters and standard image classification benchmarks (see Table~\ref{dp_sat_acc}). Furthermore, our analysis indicates that applying the generalization technique significantly enhances model accuracy (see Figure~\ref{fig:gt_accuracy}). Specifically, for private learning (i.e., $(8, 10^{-5})$-DP) trained over CIFAR-10, the accuracy improves by 31.64\% (from 49.47 to 81.11). For CIFAR-100 (using a 28-layer Wide-ResNet model pre-trained on the ImageNet), the improvement is 42.4\% (from 41.4 to 83.8). In non-private learning for CIFAR-10, there is an improvement of 21.17\% (from 71.13 to 92.3). For CIFAR-100, the enhancement is 22.07\% (from 67.62 to 91.09).

\item While prior research and ours demonstrate that generalization techniques enhance the privacy-utility balance, their impact on model bias remains uncharted. To investigate this, we employ datasets with synthetic bias,  such as CIFAR-10S~\cite{wang2020towards} and CIFAR-100S, and measure bias amplification in our private and non-private models. Our results reveal that each generalization technique, as well as their collective application, amplifies model bias (see Figure~\ref{fig:bias_accuracy}), even though they improve model accuracy (see Table~\ref{cifar_10s_bias}). In particular, in non-private models trained over CIFAR-10S, the model bias escalates by a factor of 9 (from  0.01 to 0.09), while in private models for CIFAR-10S, the model bias increases nearly by a factor of 4 (from 0.04 to 0.15). This trend of increased model bias, which is also observable for CIFAR-100S, manifests a potential compromise in model fairness. Furthermore, our results reveal that the greater the bias in the training set, the more drop in accuracy (see Figure~\ref{fig:accuracy_cifar10} and Figure~\ref{fig:accuracy_cifar100}) and increase in model bias (see Figure~\ref{fig:bias_cifar10} and Figure~\ref{fig:bias_cifar100}).

\item  We employ MIAs as a standard tool for assessing the privacy risks~\cite{de2022unlocking,song2021systematic} of our ML models while applying the generalization techniques. To conduct MIAs, we use the TensorFlow Privacy library~\cite{tfplib}, which assesses the privacy risk via the MIA AUC metric. We show that MIAs that require training multiple shadow models are more effective in both private and non-private learning settings, with biased training data (see Figure~\ref{fig:all_mem_acc_cifarS}) or without it (see Figure~\ref{fig:all_mem_acc_cifar}). Moreover, our findings show a direct link between the level of bias in the training data and an elevated MIA AUC (see Figure~\ref{fig:attack_auc_cifar10} and Figure~\ref{fig:attack_auc_cifar100}). This suggests that training biased data can make an individual's presence in the dataset more detectable. Upon implementing all generalization techniques to achieve optimal accuracy, the private model (under $(8, 10^{-5})$-DP), trained on the unbiased CIFAR-10 and CIFAR-100, experiences a decrease in MIA AUC by 0.12 and 0.11, respectively (see Table~\ref{cifar-privacy}). Conversely, this decline is limited to 0.03 for the private model trained on the biased CIFAR-10S and CIFAR-100S (see Table~\ref{cifarS-privacy}). This pattern shows that training with biased data significantly affects DP's effectiveness.

\item In the absence of a unified metric capable of integrating utility, privacy, and fairness, we take a modest step forward and introduce \textit{Harmonic Score (HS)}. It simplifies complex evaluations by calculating the harmonic mean of accuracy, privacy (measured by MIA AUC), and fairness (adjusted by bias). This score emphasizes balanced performance across these dimensions, ensuring that poor outcomes in any one aspect significantly impact the final score. For instance, as described in Figure~\ref{fig:abe}, the application of DP reduces the effectiveness of MIAs, but this comes at the cost of the model's accuracy and bias. However, the impact on HS differs across datasets. For CIFAR-100S, the addition of DP lowers HS, indicating that the trade-offs in accuracy and fairness outweigh the privacy gains, leading to a less balanced model. In contrast, for CIFAR-10, the DP-enabled model achieves a slightly higher HS, suggesting that the reduction in MIA risk (privacy gain) positively impacts the balance among the three metrics, even with trade-offs in accuracy and fairness. Furthermore, both a reduction in the bias of the training data and an increase in the privacy budget provide a better balance between privacy, utility, and fairness (see Figure~\ref{fig:overall_abe_cifar10} and Figure~\ref{fig:overall_abe_cifar100}).

\item We extend our analysis to the CelebA dataset~\cite{liu2015deep}, which captures real-world biases, such as how facial attributes correlate with gender. This analysis confirms the trends observed in our earlier experiments with CIFAR-10S and CIFAR-100S. Incorporating generalization techniques in non-private and private ( $(8, 10^{-5})$-DP) models improves model accuracy while increasing gender bias in predictions (see Table~\ref{tab:celebA}). Additionally, the results demonstrate that attributes with higher gender imbalance in the training data experience greater declines in accuracy (see Figure~\ref{fig:celebA}) and elevated MIA AUC (see Table~\ref{tab:celebA}). This validation supports our earlier conclusions about the complex interactions between accuracy, privacy, and fairness.

\item Carlini et al.~\cite{carlini2022privacy} show that eliminating the layer of outlier samples subsequently exposes another layer of samples, previously considered safe, to the same vulnerability, a phenomenon termed the Onion Effect. Our study broadens the work of Carlini et al. by analyzing models trained with and without DP on both unbiased (CIFAR-10, CIFAR-100) and biased (CIFAR-10S, CIFAR-100S) datasets, assessing MIA AUC, accuracy, bias, and HS, and deepening the investigation into the Onion Effect through the removal of three layers of outliers. This removal impacts our findings in several ways. First, it decreases accuracy regardless of its application to private or non-private learning (see Figure~\ref{fig:accuracy_onion_effect_cifar10} and Figure~\ref{fig:accuracy_onion_effect_cifar100}). Moreover, it shows a continuous exposure to MIAs (see Figure~\ref{fig:attack_acc_onion_effect_cifar10} and Figure~\ref{fig:attack_acc_onion_effect_cifar100}). Additionally, it amplifies model bias in both non-private and private learning models (see Figure~\ref{fig:bias_onion_effect_cifar10} and Figure~\ref{fig:bias_onion_effect_cifar100}). Lastly, it leads to a discordant imbalance among accuracy, privacy, and bias in non-private learning contexts (see Figure~\ref{fig:abe_onion_effect_cifar10} and Figure~\ref{fig:abe_onion_effect_cifar100}).

\end{itemize}

\section{Background}
\label{sec:background}
This section provides an overview of the DP definition and discusses a succinct overview of the generalization techniques and MIAs to offer a clearer understanding of their implications in our study.

\subsection{Differential Privacy (DP)}
DP ensures the privacy of individual contributions in statistical databases by asserting that removing or adding an individual's data does not significantly affect the outcome of any analysis. Introduced by Dwork et al.~\cite{dwork2014algorithmic}:
\begin{definition}
A mechanism $M$ satisfies \((\epsilon, \delta)\)-DP if for all datasets $D_1$ and $D_2$ differing on at most one element, and for all $S \subseteq Range(M)$, it holds that
\[
Pr[M(D_1) \in S] \leq \exp(\epsilon) \cdot Pr[M(D_2) \in S] + \delta.
\]
\end{definition}
Here, $\epsilon$ referred to as the privacy loss, where a smaller value indicates higher privacy. $\delta$ represents a small probability that the privacy loss may be exceeded. 


DP establishes a framework for developing private ML models, highlighting the role of DP-SGD as a pivotal technique in this context~\cite{abadi2016deep}. In DP-SGD, noise is added to the gradients during each update step to mask the influence of individual data points, thus ensuring that the training process remains differentially private. The effectiveness of DP-SGD relies on the privacy accountant—a numerical algorithm that calculates precise upper bounds on cumulative privacy loss, which is called privacy budget~\cite{abadi2016deep}. The privacy accountant tracks how privacy loss accumulates over multiple training iterations, ensuring that the total privacy loss remains within a specified budget. In our study, we employ the accounting method for DP-SGD introduced by Mironov et al.~\cite{mironov2019r}, which is available in the TensorFlow Privacy library~\cite{tfplib}. De et al.~\cite{de2022unlocking} utilize this approach, combining privacy accounting with meticulous hyper-parameter optimization to enhance the accuracy of over-parameterized models. This results in a refined trade-off between privacy and utility, achieving state-of-the-art outcomes. In the subsequent sections, we elaborate on the generalization techniques proposed by De et al.~\cite{de2022unlocking} and DP-SAT by Park et al.~\cite{park2023differentially}, and detail their implications in our study.

\subsection{Generalization Techniques}
\textbf{Group Normalization (GN).} Following recent studies~\cite{de2022unlocking, kurakin2022toward, luo2021scalable}, we replace Batch Normalization (BN) layers by GN layers. This modification is important because DP-SGD requires independent gradients evaluated on different training examples. This fails to include any method that enables communication between training examples, such as BN. GN, on the other hand, divides the channels of the hidden activations of a single image into groups and normalizes these activations within each group independently. This maintains the independence between gradients evaluated on different examples. Following De et al.~\cite{de2020batch}, we place the GN layers on the residual branch of the network to recover the benefits of BN for training deep networks. 

\textbf{Optimal Batch Size (OBS).}
Previous studies~\cite{de2022unlocking,anil2021large,dormann2021not} have noted that using larger batch size can notably boost the privacy-utility balance in DP-SGD. On the other hand, in non-private models using SGD, the batch size is typically smaller (e.g., 8, 16, 32, 64) to achieve higher accuracy~\cite{tan2019efficientnet, du2021efficient}. 

\textbf{Weight Standardization (WS).}
Several studies~\cite{kolesnikov2020big,qiao1903micro,richemond2020byol} have shown that using WS combined with GN can be an effective replacement for BN in non-private training, especially when training with large BS. Adopting the approach of~\cite{de2022unlocking}, we use this technique for all convolutional layers to normalize the rows of the weight matrix for each convolution and demonstrate its advantage in private learning.

\textbf{Augmentation Multiplicity (AM).} As in~\cite{de2022unlocking}, we use multiple augmentations for each sample in DP-SGD updates to regain the advantages of data augmentation in private training, and instead of calculating a clipped gradient for every augmented image—which would increase privacy costs—we average the gradients from various augmentations of a single training sample before gradient clipping. This method does not impose extra privacy expenses.

\textbf{Parameter Averaging (PA).}
The PA technique~\cite{polyak1992acceleration} leverages the stability of parameters over training iterations to enhance model generalization by averaging parameters across multiple steps. This approach helps smooth parameter updates, leading to more robust model performance. The privacy analysis of DP-SGD assumes that revealing training parameters does not breach privacy; therefore, PA does not result in additional privacy concerns. Following~\cite{de2022unlocking}, we adopt an exponential moving average for PA, which continuously updates a weighted average of the parameters during training. This method improves accuracy on both training and validation data by reducing the variance of parameter updates, providing a stable optimization path for improved model performance.

\textbf{Sharpness-Aware Training (SAT).} 
Sharpness-aware minimization (SAM) targets flat minima to mitigate the issue of sharp minima in the loss landscape~\cite{foret2021sharpnessaware}. This characteristic of sharp minima is recognized as a limitation of SGD in yielding generalized models, particularly in the context of over-parameterized models. 

Unlike SGD, which computes the gradients of the loss function relative to the parameters and then guides the parameter updates in a single descent step, SAM introduces a two-step optimization. This method initially perturbs the parameters within a certain radius in the ascent step to evaluate the sensitivity of the loss function (a measure of the landscape's sharpness) and subsequently steers the perturbed parameters toward flatter regions of the loss landscape in the descent step. 

However, SAM's two-step optimization may negatively impact the privacy budget and computational time of DP-SAM~\cite{du2021efficient,park2023differentially}. Specifically, Park et al.~\cite{park2023differentially} prove that DP-SAM requires twice the privacy budget than that of DP-SGD and requires more computational time. This is because DP-SAM employs the training samples within the same mini-batch twice, i.e., it needs to inject noise into both the gradients of the current parameters and the perturbed parameters to ensure the privacy of both ascent and descent steps. To mitigate this challenge, Park et al. propose DP-SAT, which can improve performance without additional privacy or computational burden. Their idea is to reuse the perturbed gradient of the previous step to steer the direction of updated parameters at the current step. Our study distinctively examines the impact of generalization techniques introduced by~\cite{de2022unlocking} and the SAT optimizer on the fairness and accuracy of ML models, with and without DP, addressing a gap not explored in prior research.

\begin{table*}
\small
\centering
\caption{Hyper-parameters for CIFAR-10(S), CIFAR-100(S), and CelebA datasets with and without private learning}
\label{setting}
\begin{tabular}{|c|c|c|c|c|c|c|c|c|c|c|c|c|} 
\hline
\multirow{2}{*}{Hyper-parameter} & \multicolumn{9}{c|}{With DP}                                                                                                               & \multicolumn{3}{c|}{Without DP}                                                                              \\ 
\cline{2-13}
                                 & \multicolumn{4}{c|}{CIFAR-10(S)}              & \multicolumn{4}{c|}{CIFAR-100(S)}             & \multicolumn{1}{c|}{CelebA}              & \multirow{3}{*}{CIFAR-10(S)} & \multicolumn{1}{c|}{\multirow{3}{*}{CIFAR-100(S)}} & \multicolumn{1}{c|}{\multirow{3}{*}{CelebA}}  \\ 
\cline{1-10}
$\epsilon$                       & 1         & 2         & 4         & 8         & 1         & 2         & 4         & 8         & 8                                          &                              & \multicolumn{1}{c|}{}                              &                          \\ 
\cline{1-10}
$\delta$                         & $10^{-5}$ & $10^{-5}$ & $10^{-5}$ & $10^{-5}$ & $10^{-5}$ & $10^{-5}$ & $10^{-5}$ & $10^{-5}$ & $2.02 \times 10^{-5}$ &                              & \multicolumn{1}{c|}{}                              &                          \\ 
\hline
Learning rate                    & 2         & 2         & 4         & 4         & 1         & 1         & 1         & 1         & 1                                           & 0.01                         & 0.01                                               & $10^{-4}$         \\ 
\hline
Noise multiplier                 & 10        & 6         & 4         & 3         & 21.1      & 15.8      & 12        & 9.4       & 6                                           & -                            & -                                                  & -                        \\ 
\hline
GN                               & 16        & 16        & 16        & 16        & 16        & 16        & 16        & 16        & 16                                          & 1                            & 1                                                  & 16                       \\ 
\hline
OBS                              & 4096      & 4096      & 4096      & 4096      & 16348     & 16348     & 16348     & 16348     & 4096                                        & 32                           & 64                                                 & 64                       \\ 
\hline
AM                               & 16        & 16        & 16        & 16        & 16        & 16        & 16        & 16        & -                                           & 16                           & 16                                                 & -                        \\ 
\hline
Radius (SAT parameter)           & 0.05      & 0.05      & 0.05      & 0.05      & 0.03      & 0.03      & 0.03      & 0.03      & 0.03                                        & 0.05                         & 0.03                                               & 0.03                     \\
\hline
\end{tabular}
\end{table*}

\subsection{Membership Inference Attacks (MIAs)}
An MIA targets a built ML model to deduce membership of individual training samples. This can have privacy implications, especially when the training data includes sensitive or personal information~\cite{shokri2017Membership,salem2018ml}. MIAs can be conducted using two main strategies differing in their need for training shadow models. 

First, as discussed by Shokri et al.~\cite{shokri2017Membership}, the adversary initially trains multiple shadow models to simulate the target model, assuming that the target model is a
black-box API. Then, based on shadow models’ outputs on their own training and test examples, the adversary obtains a labeled (member vs non-member) dataset and finally trains multiple neural network classifiers, one for each class label, to perform MIAs against the target model. This approach capitalizes on the similarity between the shadow models and the target model to infer membership.

The second approach, by Salem et al.~\cite{salem2018ml}, relies on training one shadow model to distinguish between member and non-member data points. This approach is computationally more efficient as it avoids the overhead of training multiple shadow models.

To conduct MIAs, we use the TensorFlow Privacy library~\cite{tfplib} in which the focus is on MIAs against black-box models, where the adversary can only observe the model’s output but not its parameters. We employ four different MIAs implemented in this library: MultiLayered Perceptron (MLP), Random Forest (RF), Threshold, and Threshold Entropy attacks. The first two options require training one shadow model; the last two do not require training any shadow model and leverage statistical measures, such as maximum confidence score and entropy, applied to the target model's results. The Threshold attack uses a simple decision rule based on the confidence score, while the Threshold Entropy attack uses the entropy of the confidence scores to make membership decisions.

Figure~\ref{fig:all_mem_acc_cifarS} and Figure~\ref{fig:all_mem_acc_cifar} highlight the superior efficacy of the MLP attack over the others across models trained on various datasets. The model trained using DP on CIFAR-100S is an exception, where the RF attack outperforms others. Consequently, we report MIA AUC of the MLP attack across all experiments except for the DP-trained model on CIFAR-100S, where we specifically report MIA AUC of the RF attack.

\section{Methodology and Experimental Setup}
This section provides the methodologies and metrics employed in our experiments. It first discusses our selection and manipulation of datasets. It then explains the architectural choices and training settings for our ML models. Additionally, it introduces the MIA AUC and bias metrics from the literature, which are used for assessing privacy and fairness. It also describes our HS metric, an approach that simultaneously evaluates model accuracy, fairness, and privacy.

\subsection{Datasets}
Our analysis examines bias in both controlled, simplified settings and more complex, real-world scenarios. In the simplified setting, we use the CIFAR-10 and CIFAR-100 datasets for training with unbiased data and introduce bias through their skewed counterparts, CIFAR-10S~\cite{wang2020towards} and CIFAR-100S.

\textbf{Synthetic Bias with CIFAR-10S and CIFAR-100S.} CIFAR-10S is created by converting a subset of images to grayscale, maintaining a 95\% to 5\% ratio between color and grayscale images per class. Specifically, five classes are predominantly color (95\%), while the other five are mainly grayscale (95\%). Despite this skew at the class level, the overall distribution between color and grayscale images remains balanced. For evaluation, we use two test sets: a color-only version (COLOR) and a grayscale-only version (GRAY), each assessed independently for the 10-class classification. CIFAR-100S follows the same structure but extends to the 100 classes of CIFAR-100. Unless otherwise specified, the bias ratio is set at 95\% to 5\%, though we also explore a 75\% to 25\% ratio to understand the impact of varying degrees of skewness in the data distribution.

\textbf{Real-World Bias with CelebA.} To extend our analysis to a realistic setting, we select the CelebA dataset~\cite{liu2015deep}, which naturally exhibits imbalances in attribute distribution across genders. We focus on the Aligned\&Cropped subset, which contains images with 39 facial attributes, allowing us to study how attributes like "smiling" correlate with gender. Among the attributes, 21 are more commonly associated with women, while 18 are more frequent in men, showing an average gender bias of 80.0\% when an attribute is present. We exclude the “Male” attribute and focus on attributes that have a sufficient number of validation and test images, ultimately analyzing 34 attributes. This dataset allows us to validate our findings from the controlled setting and assess how models handle the more nuanced biases that arise in real-world data.

\subsection{Metrics}
\textbf{Bias Metric.} To measure bias amplification in our models, we employ two distinct metrics introduced in~\cite{wang2020towards} suited to the nature of the biased datasets used in our study.
\begin{itemize}
    \item \textbf{Synthetic Bias.} In the context of CIFAR-10S and CIFAR-100S, we use a bias metric that calculates the mean bias in model predictions across classes:
    
    \begin{equation}
    \text{Bias} = \frac{1}{|C|} \sum_{c \in C} \left(\frac{\max(\mathrm{Gr}_c, \mathrm{Col}_c)}{\mathrm{Gr}_c + \mathrm{Col}_c} - 0.5\right),
    \label{eq:bias_metric1}
    \end{equation}   
    where \( \mathrm{Gr}_c \) is the number of grayscale test set examples predicted as class \( c \), and \( \mathrm{Col}_c \) is the same for color. Here, \( C \) represents the set of all classes. The test set is evenly distributed across the domains \{Gr, Col\}, ensuring that the average accuracy directly reflects the model's learned bias towards one of the domains in the model’s predictions. This metric helps identify the extent to which the model's predictions favor either grayscale or color images, reflecting how spurious correlations learned during training influence the model’s outcomes.

    \item \textbf{Real-World Bias.} For each attribute in the CelebA dataset, we calculate model bias to assess how much the model’s predictions reflect or amplify inherent gender imbalances:
    \begin{equation}
    \text{Bias} = \frac{P_w}{P_w + P_m} - \frac{N_w}{N_w + N_m},
    \label{eq:bias_metric2}
    \end{equation}
    where \( P_w \) and \( P_m \) are the numbers of positive classifications for women and men, respectively, and \( N_w \) and \( N_m \) are the actual counts of images with the attribute present for women and men. If an attribute is more common among women (e.g., "smiling"), a positive value indicates that the model is amplifying the gender imbalance by predicting more women as having the attribute than the original distribution suggests. A negative value implies a reduction in bias. This metric helps assess how the model's predictions align with or deviate from the underlying distribution of attributes across genders, aiming for more balanced outcomes.
\end{itemize}

\textbf{MIA AUC.} Following~\cite{salem2018ml,watson2021importance,de2022unlocking}, we use the area under the curve (AUC) metric to measure the effectiveness of MIAs on our ML models. It represents the area under the receiver operating characteristic (ROC) curve, which plots the true positive rate against the false positive rate. High AUC values demonstrate a model's vulnerability by showing its ability to distinguish between training dataset members and non-members, thus revealing susceptibility to MIAs. In contrast, AUC values around 0.5 imply the model's robustness against such attacks.

\begin{table*}
\centering
\caption{An ablation study on the effect of architectural modifications and changes to the training pipeline for models trained on CIFAR-10 and CIFAR-100. DP-CIFAR-10 and DP-CIFAR-100 refer to these datasets used in the private setting under $(8, 10^{-5})$-DP. We report model accuracy (acc) and its standard deviation (in \textcolor[rgb]{0.502,0.502,0.502}{gray}) on official test sets, and the highest MIA AUC (AUC) among four attacks belonging to MLP is presented. Additionally, the generalization gap (GGap) is also reported.}
\label{cifar-privacy}
\begin{tabular}{|l|c|c|c|c|c|c|c|c|c|c|c|c|} 
\hline
\multirow{2}{*}{} & \multicolumn{3}{c|}{DP-CIFAR-10}                           & \multicolumn{3}{c|}{CIFAR-10}                              & \multicolumn{3}{c|}{~DP-CIFAR-100}                         & \multicolumn{3}{c|}{CIFAR-100}                               \\ 
\cline{2-13}
                  & acc                                             & AUC&GGap & acc                                             & AUC&GGap & acc                               & AUC &GGap& acc                            & AUC&GGap  \\ 
\hline
Baseline~         & 49.47 \textcolor[rgb]{0.502,0.502,0.502}{(1.3)} & 0.55  &  3.76 & 71.13 \textcolor[rgb]{0.502,0.502,0.502}{(0.9)} & 0.55   & 1.39 & 41.4 \textcolor[rgb]{0.502,0.502,0.502}{(1.3)}  & 0.57  & 18.32  & 67.62 \textcolor[rgb]{0.502,0.502,0.502}{(0.6)}  & 0.58  &   11.13 \\ 
\hline
+ GN              & 54.96 \textcolor[rgb]{0.502,0.502,0.502}{(0.6)} & 0.55   & 2.48 & 76.08 \textcolor[rgb]{0.502,0.502,0.502}{(1.1)} & 0.55  & 0.96  & 48.6 \textcolor[rgb]{0.502,0.502,0.502}{(0.6)}  & 0.56 &  16.59  & 73.05 \textcolor[rgb]{0.502,0.502,0.502}{(0.4)}  & 0.59  &  9.23  \\ 
\hline
+ OBS              & 73.76 \textcolor[rgb]{0.502,0.502,0.502}{(0.8)} & 0.55  & 2.24  & 83.7 \textcolor[rgb]{0.502,0.502,0.502}{(0.8)}  & 0.54 &  0.68  & 69.3 \textcolor[rgb]{0.502,0.502,0.502}{(0.8)}  & 0.55  &  12.83 & 81.5 \textcolor[rgb]{0.502,0.502,0.502}{(0.3)}   & 0.64   & 6.84  \\ 
\hline
+ WS              & 74.37 \textcolor[rgb]{0.502,0.502,0.502}{(0.6)} & 0.56   & 2.17 & 84.3 \textcolor[rgb]{0.502,0.502,0.502}{(0.6)}  & 0.67  & 0.63  & 72.8 \textcolor[rgb]{0.502,0.502,0.502}{(0.6)}  & 0.54  & 12.31  & 82.7 \textcolor[rgb]{0.502,0.502,0.502}{(0.3)}   & 0.65  &  6.46  \\ 
\hline
+ AM              & 78.14 \textcolor[rgb]{0.502,0.502,0.502}{(0.5)} & 0.54  & 2.11  & 91.58 \textcolor[rgb]{0.502,0.502,0.502}{(0.3)} & 0.63  &  0.41 & 80.7 \textcolor[rgb]{0.502,0.502,0.502}{(0.5)}  & 0.55  &  11.69 & 89.4 \textcolor[rgb]{0.502,0.502,0.502}{ (0.2)}   & 0.64 &  5.38  \\ 
\hline
+ PA              & 79.06 \textcolor[rgb]{0.502,0.502,0.502}{(0.7)} & 0.53  & 2.05  & 91.64 \textcolor[rgb]{0.502,0.502,0.502}{(0.2)} & 0.63  & 0.35  & 83.1 \textcolor[rgb]{0.502,0.502,0.502}{(0.3)}  & 0.55 &  10.19  & 90.37 \textcolor[rgb]{0.502,0.502,0.502}{(0.1)}  & 0.66   &  5.13 \\ 
\hline
+ SAT             & 81.11 \textcolor[rgb]{0.502,0.502,0.502}{(0.3)} & 0.53  &  1.94 & 92.3 \textcolor[rgb]{0.502,0.502,0.502}{(0.2)}  & 0.65  &  0.26 & 83.8 \textcolor[rgb]{0.502,0.502,0.502}{(0.2)}  & 0.56  & 10.08  & 91.09 \textcolor[rgb]{0.502,0.502,0.502}{(0.1)}  & 0.66  &  4.88  \\
\hline
\end{tabular}
\end{table*}

\textbf{Harmonic Score (HS).}  
We introduce a scoring metric based on the harmonic mean to evaluate models in terms of privacy (measured by MIA AUC), fairness (adjusted by bias), and accuracy. To ensure all aspects are considered equally, we map values to the \([0, 1]\) range:

\begin{itemize}
    \item \(\text{Bias}\) is measured using Eq.~\ref{eq:bias_metric1}, resulting values within the range \([0, 0.5]\), where \(0\) represents no bias (perfect fairness) and \(0.5\) represents maximum bias. We scale this to \([0, 1]\), resulting in \(\text{Bias}_{\text{scaled}}\). We then use \(1 - \text{Bias}_{\text{scaled}}\), where values closer to 1 indicate lower bias, and values closer to 0 represent higher bias.
    
    \item \(\text{AUC}\) for MIA falls in the range \([0.5, 1]\), with \(0.5\) indicating random guessing (no successful attack) and \(1\) representing a successful attack. We map this to \([0, 1]\), yielding \(\text{AUC}_{\text{scaled}}\), and then use \(1 - \text{AUC}_{\text{scaled}}\), where values closer to 1 indicate stronger privacy, and values closer to 0 indicate weaker privacy.
    
    \item Accuracy is already in the \([0, 1]\) range, where higher values indicate better model performance.
\end{itemize}

The Harmonic Score (HS) is calculated using the following formula:

\begin{equation}
    \text{HS} = \frac{3}{\frac{1}{\text{Accuracy}} + \frac{1}{1 - \text{AUC}_{\text{scaled}}} + \frac{1}{1 - \text{Bias}_{\text{scaled}}}}
\end{equation}

HS is heavily influenced by the lowest value among the three aspects—accuracy, \(1 - \text{Bias}_{\text{scaled}}\), and \(1 - \text{AUC}_{\text{scaled}}\)—ensuring that poor performance in one dimension significantly reduces the overall score. High values for each of these aspects lead to a higher harmonic mean, indicating balanced and strong performance across all three dimensions. HS ranges from \((0, 1]\), with higher values (closer to 1) indicating better overall performance, while lower values (closer to 0) reflecting poorer performance in at least one aspect.

For example, consider the following two cases:
\begin{itemize}
    \item \textbf{Case 1}: (accuracy = 0.5, \(1 - \text{bias} = 0.5\), \(1 - \text{AUC} = 0.5\)) \\
    \[
    \text{HS} = \frac{3}{\frac{1}{0.5} + \frac{1}{0.5} + \frac{1}{0.5}} = 0.5
    \]

    \item \textbf{Case 2}: (accuracy = 0.1, \(1 - \text{bias} = 0.5\), \(1 - \text{AUC} = 0.9\)) \\
    \[
    \text{HS} = \frac{3}{\frac{1}{0.1} + \frac{1}{0.9} + \frac{1}{0.5}} \approx 0.136
    \]
\end{itemize}

Both cases have the same arithmetic mean value of \(\frac{0.5 + 0.5 + 0.5}{3} = 0.5\). However, HS for Case 2 is significantly lower due to its poor accuracy performance. This illustrates how the harmonic mean is more sensitive to low-performing aspects, emphasizing the importance of balanced performance across all three dimensions.

\subsection{Implementation Details}
We train our deep ML model using the Wide-ResNet (WRN) architecture, specifically WRN-16-4 for CIFAR-10(S) and WRN-28-10 for CIFAR-100(S) and CelebA. The baseline model is trained from scratch on CIFAR-10(S) but pre-trained on ImageNet~\cite{russakovsky2015imagenet} for CIFAR-100(S) and CelebA experiments.

We use the parameters relevant to the generalization techniques and our private and non-private settings, as detailed in Table~\ref{setting}. For experiments on CIFAR-10 and CIFAR-100 datasets, we follow the setting for private models as used in~\cite{de2022unlocking, park2023differentially}. For non-private scenarios, we determined the optimal parameters through experimental exploration. This included testing various batch sizes (8, 16, 32, 64, 128), setting different learning rates (0.01, 0.1), evaluating multiple values (1, 2, 4, 8, 16) for GN and AM, and assessing different values (0.03, 0.04, 0.05) for radius. The CelebA's training is conducted using binary cross-entropy loss with logits. We employ early stopping, as ~\cite{song2021systematic}, to prevent overfitting in non-private models. We perform five independent runs with different seeds on each experiment and report their median, and utilize a server with 8 RTX 4090 GPUs for the computational requirement. Our implementation is publicly available at \url{https://anonymous.4open.science/r/PriFa_ML-D04A}.

\section{Experimental Results}
In this section, we empirically evaluate the impact of various generalization techniques on privacy-utility and fairness-utility trade-offs, as well as their collective effects on privacy, utility, and fairness in ML models. We systematically integrate these techniques into the training process and analyze their outcomes across different datasets and settings. Additionally, we address the Onion Effect, where removing layers of outlier samples reveals new vulnerabilities, to gain insights into model robustness against privacy attacks. For a detailed breakdown of each generalization technique's individual effect, please refer to the Supplementary material, Table~\ref{cifar-privacy-gt}.

\subsection{Privacy-Utility Trade-off}
We first assess the influence of the generalization techniques on the privacy-utility trade-off (addressing \textbf{Q1}). For this assessment, we train our models on CIFAR-10, CIFAR-10S, CIFAR-100, and CIFAR-100S. We cumulatively integrate each generalization technique into the SGD training process as proposed by~\cite{de2022unlocking}, in settings with and without DP.  As the culmination of our technique integration, we replace SGD with SAT as the final generalization technique in our sequence.  After each integration, we subject the model to the four listed MIA attacks. This approach allows us to systematically evaluate the cumulative impact of these techniques and understand their effect on model accuracy and privacy. We provide a deeper analysis of how these techniques impact privacy leakage later in this section. We also compare DP-SGD and DP-SAT for different privacy budgets and datasets to evaluate their accuracy when both leverage the generalization techniques.

\begin{table*}
\caption{An ablation study on the effect of architectural modifications and changes to the training pipeline for models trained on CIFAR-10S and CIFAR-100S. DP-CIFAR-10S and DP-CIFAR-100S refer to these datasets in the private setting under $(8, 10^{-5})$-DP, while CIFAR-100S and CIFAR-10S mention the non-private setting (without DP). We report model accuracy (acc) for Color (C) and Gray (G) test sets and their standard deviation (in \textcolor[rgb]{0.502,0.502,0.502}{gray}), and the highest MIA AUC (AUC) among the four attacks is reported.}
\centering
\label{cifarS-privacy}
\begin{tabular}{|l|c|c|c|c|c|c|c|c|} 
\hline
\multirow{2}{*}{} & \multicolumn{2}{c|}{DP-CIFAR-10S}           & \multicolumn{2}{c|}{CIFAR-10S}              & \multicolumn{2}{c|}{~DP-CIFAR-100S}                                                                                 & \multicolumn{2}{c|}{CIFAR-100S}                                                                                      \\ 
\cline{2-9}
                  & acc (C\textbar{}G)               & AUC & acc (C\textbar{}G)               & AUC & acc (C\textbar{}G)                                                                                       & AUC & acc (C\textbar{}G)                                                                                       & AUC  \\ 
\hline
Baseline~         & 53.8 \textcolor[rgb]{0.502,0.502,0.502}{(1.8)} \textbar{} 53.6 \textcolor[rgb]{0.502,0.502,0.502}{(1.7)} & 0.85      & 67.3 \textcolor[rgb]{0.502,0.502,0.502}{(0.4)} \textbar{} 65.8 \textcolor[rgb]{0.502,0.502,0.502}{0.3)} & 0.85     & 19.4 \textcolor[rgb]{0.502,0.502,0.502}{(0.6)} \textbar{} 19.2 \textcolor[rgb]{0.502,0.502,0.502}{(0.6)} & 0.65     & 57.9 \textcolor[rgb]{0.502,0.502,0.502}{(0.2)} \textbar{} 57.6 \textcolor[rgb]{0.502,0.502,0.502}{(0.2)} & 0.76      \\ 
\hline
+ GN              & 56.3 \textcolor[rgb]{0.502,0.502,0.502}{(0.5)} \textbar{} 56.1 \textcolor[rgb]{0.502,0.502,0.502}{(0.5)} & 0.85     & 66.6 \textcolor[rgb]{0.502,0.502,0.502}{(0.2)} \textbar{} 66.8 \textcolor[rgb]{0.502,0.502,0.502}{(0.2)} & 0.83     & 23.1 \textcolor[rgb]{0.502,0.502,0.502}{(0.4)} \textbar{} 22.8 \textcolor[rgb]{0.502,0.502,0.502}{(0.4)} & 0.65     & 59.7 \textcolor[rgb]{0.502,0.502,0.502}{(0.2)} \textbar{} 58.9 \textcolor[rgb]{0.502,0.502,0.502}{(0.3)} & 0.75      \\ 
\hline
+ OBS              & 60.7 \textcolor[rgb]{0.502,0.502,0.502}{(0.3)} \textbar{} 60.5 \textcolor[rgb]{0.502,0.502,0.502}{(0.3)} & 0.84     & 69.5 \textcolor[rgb]{0.502,0.502,0.502}{(0.3)} \textbar{} 67.1 \textcolor[rgb]{0.502,0.502,0.502}{(0.3)} & 0.83     & 28.4 \textcolor[rgb]{0.502,0.502,0.502}{(0.2)} \textbar{} 28.5 \textcolor[rgb]{0.502,0.502,0.502}{(0.2)} & 0.66     & 60.9 \textcolor[rgb]{0.502,0.502,0.502}{(0.1)} \textbar{} 60.5 \textcolor[rgb]{0.502,0.502,0.502}{(0.1)} & 0.75      \\ 
\hline
+ WS              & 61.4 \textcolor[rgb]{0.502,0.502,0.502}{(0.6)} \textbar{} 61.1 \textcolor[rgb]{0.502,0.502,0.502}{(0.4)} & 0.81     & 71.7 \textcolor[rgb]{0.502,0.502,0.502}{(0.2)} \textbar{} 69.5 \textcolor[rgb]{0.502,0.502,0.502}{(0.5)} & 0.83     & 29.8 \textcolor[rgb]{0.502,0.502,0.502}{(0.1)} \textbar{} 30.1 \textcolor[rgb]{0.502,0.502,0.502}{(0.1)} & 0.65     & 63.3 \textcolor[rgb]{0.502,0.502,0.502}{(0.1)} \textbar{} 63.1 \textcolor[rgb]{0.502,0.502,0.502}{(0.1)} & 0.73      \\ 
\hline
+ AM              & 62.6 \textcolor[rgb]{0.502,0.502,0.502}{(0.2)} \textbar{} 62.6 \textcolor[rgb]{0.502,0.502,0.502}{(0.3)} & 0.81     & 75.4 \textcolor[rgb]{0.502,0.502,0.502}{(0.1)} \textbar{} 72.9 \textcolor[rgb]{0.502,0.502,0.502}{(0.2)} & 0.85     & 31.6 \textcolor[rgb]{0.502,0.502,0.502}{(0.3)} \textbar{} 31.8 \textcolor[rgb]{0.502,0.502,0.502}{(0.4)} & 0.65     & 69.8 \textcolor[rgb]{0.502,0.502,0.502}{(0.3)} \textbar{} 69.5 \textcolor[rgb]{0.502,0.502,0.502}{(0.1)} & 0.71      \\ 
\hline
+ PA              & 65.2 \textcolor[rgb]{0.502,0.502,0.502}{(0.1)} \textbar{} 65.1 \textcolor[rgb]{0.502,0.502,0.502}{(0.1)} & 0.81     & 84.7 \textcolor[rgb]{0.502,0.502,0.502}{(0.1)} \textbar{} 82.4 \textcolor[rgb]{0.502,0.502,0.502}{(0.1)} & 0.84     & 36.5 \textcolor[rgb]{0.502,0.502,0.502}{(0.1)} \textbar{} 36.6 \textcolor[rgb]{0.502,0.502,0.502}{(0.1)} & 0.68     & 71.3 \textcolor[rgb]{0.502,0.502,0.502}{(0.1)} \textbar{} 71.2 \textcolor[rgb]{0.502,0.502,0.502}{(0.1)} & 0.72      \\ 
\hline
+ SAT             & 65.4 \textcolor[rgb]{0.502,0.502,0.502}{(0.1)} \textbar{} 65.2 \textcolor[rgb]{0.502,0.502,0.502}{(0.2)} & 0.82     & 86.8 \textcolor[rgb]{0.502,0.502,0.502}{(0.1)} \textbar{} 85.1 \textcolor[rgb]{0.502,0.502,0.502}{(0.1)} & 0.85     & 37.2 \textcolor[rgb]{0.502,0.502,0.502}{(0.1)} \textbar{} 37.3 \textcolor[rgb]{0.502,0.502,0.502}{(0.1)} & 0.68     & 72.1 \textcolor[rgb]{0.502,0.502,0.502}{(0.1)} \textbar{} 72.3 \textcolor[rgb]{0.502,0.502,0.502}{(0.2)} & 0.71      \\
\hline
\end{tabular}
\end{table*}

Table~\ref{cifar-privacy} presents the impact of the generalization techniques on the accuracy, MIA AUC, and generalization gap of private (i.e., $(8, 10^{-5})$-DP) and non-private models trained on the CIFAR-10 and CIFAR-100. The generalization gap (GGap) is reported as the difference between the accuracy on the validation and training sets. We first discuss the results of models trained over CIFAR-10. Notably, incorporating all generalization techniques leads to an accuracy boost of 31.64\% and 21.17\% for private and non-private baseline models, respectively. These substantial increases highlight the effectiveness of the generalization techniques in enhancing model performance. Furthermore, this approach experiences a slight decrease in MIA AUC for the private baseline model but increases MIA AUC by 0.1 for the non-private baseline model. This shows that DP fortifies the models' defense against MIAs despite the accuracy gains attributed to the generalization techniques. OBS and AM are the most effective techniques for increasing the accuracy of both private and non-private models. Their significant contributions underscore their critical role in optimizing model performance. Additionally, WS significantly increases MIA AUC in the non-private setting. This increase indicates a trade-off between accuracy improvement and privacy risk, which needs consideration in model development. DP-SAT and SAT contribute to a boost of 2.05\% and 0.66\% in the accuracy of private and non-private models, respectively. These results demonstrate the nuanced impact of the SAT method on model performance. In the context of MIA AUC, DP-SAT does not impact the private model, and SAT shows a marginal increase of 0.02 in the non-private setting. 

In Table~\ref{cifar-privacy}, we also conduct our experiments on CIFAR-100, in which applying all generalization techniques notably enhances model accuracy by 42.7\%  and 23.47\% for the private and non-private baseline models, respectively. The considerable improvements for CIFAR-100 further validate the effectiveness of these techniques across different datasets. This approach increases MIA AUC by 0.08 for the non-private baseline model while exhibiting minimal fluctuation in the private models, underscoring the protective impact of DP against MIAs in private learning. OBS and AM are the key strategies for improving accuracy across the private and non-private models. Their repeated effectiveness across different datasets highlights their importance in enhancing model utility. OBS also plays a negative role in elevating MIA AUC in the non-private context. This negative impact emphasizes the need to balance accuracy improvements with potential privacy risks. Furthermore, both SAT and DP-SAT increase accuracy by approximately 0.7\%. This consistent improvement with the SAT technique indicates its beneficial role in refining model accuracy. Finally, upon applying all generalization techniques to achieve peak accuracy, MIA AUC of the private model trained on the CIFAR-10 and CIFAR-100 experiences a decrease of 0.12 and 0.1, respectively, compared to the non-private model trained on these datasets. 

By using the early stopping technique in non-private settings, we prevent the training process from reaching the point of overfitting. In private settings, the limited privacy budget restricts the number of training iterations, which also helps prevent overfitting. The GGap values reported in Table~\ref{cifar-privacy} show that DP-enabled models, particularly those trained on CIFAR-100, exhibit a higher GGap. This increase in GGap is due to the noise introduced for privacy, which impacts accuracy. The application of generalization techniques effectively reduces GGap in both private and non-private models, though non-private models consistently maintain a lower GGap.

Table~\ref{cifarS-privacy} evaluates the impact of the generalization techniques on the accuracy and MIA AUC of models trained with and without DP over CIFAR-10S and CIFAR-100S for Color (C) and Gray (G) test sets shown by C$|$G. Implementing all generalization techniques improves the accuracy of the baseline model trained over CIFAR-10S by almost 11\%  and 20\%  for the color and gray datasets in the private and non-private models, respectively. These improvements highlight the substantial impact of generalization techniques in enhancing model performance under both private and non-private settings. For CIFAR-100S, these accuracy improvements are approximately 18\% and 14\% for the private and non-private baseline models, respectively. The consistent gains across different datasets emphasize the robustness of these techniques. Incorporating all generalization techniques into the private baseline model trained on CIFAR-10S decreases MIA AUC by 0.08. However, these techniques increase MIA AUC by 0.03 for the private baseline model trained on CIFAR-100S. MIA AUC changes are marginal in the non-private model trained on CIFAR-10S but show a decrease of 0.05 in the non-private model trained on CIFAR-100S. 

Moreover, after applying all generalization techniques to achieve peak accuracy, MIA AUC of the private models trained on CIFAR-10S and CIFAR-100S experiences a decrease of 0.03 for both datasets compared to the non-private models trained on these datasets. OBS and PA are the key techniques in accuracy improvement across private models trained on CIFAR-10S and CIFAR-100S. Among others, PA leads to the most accuracy improvement in the non-private model on CIFAR-10S, while AM does this role for the non-private model on CIFAR-100S. Furthermore, both SAT and DP-SAT increase accuracy in private and non-private models. 

\subsubsection{Detailed Analysis of the Impact of Generalization Techniques on Privacy Leakage}
We now provide a closer examination of the impact of generalization techniques on privacy leakage (measured by MIA AUC), specifically for models trained on unbiased datasets CIFAR-10 and CIFAR-100, as shown in Table~\ref{cifar-privacy}, and contextualize these results with previous studies. While Table~\ref{cifar-privacy} highlights that certain generalization techniques can lead to increased MIA AUC, this outcome aligns with findings in prior research~\cite{long2018understanding, tan2023blessing}, which indicate that generalization techniques may be less effective in safeguarding vulnerable samples. Similarly, in the context of large language models,~\cite{carlini2021extracting} demonstrates that memorization of vulnerable samples can occur independently of overfitting. Here, memorization is defined following Feldman~\cite{feldman2020does} as unintended retention of specific samples, where the model’s prediction probability for a training sample changes significantly if the sample is removed from the dataset. 

\begin{figure}
    \centering
      \begin{subfigure}{0.48\linewidth}
        \includegraphics[width=\textwidth]{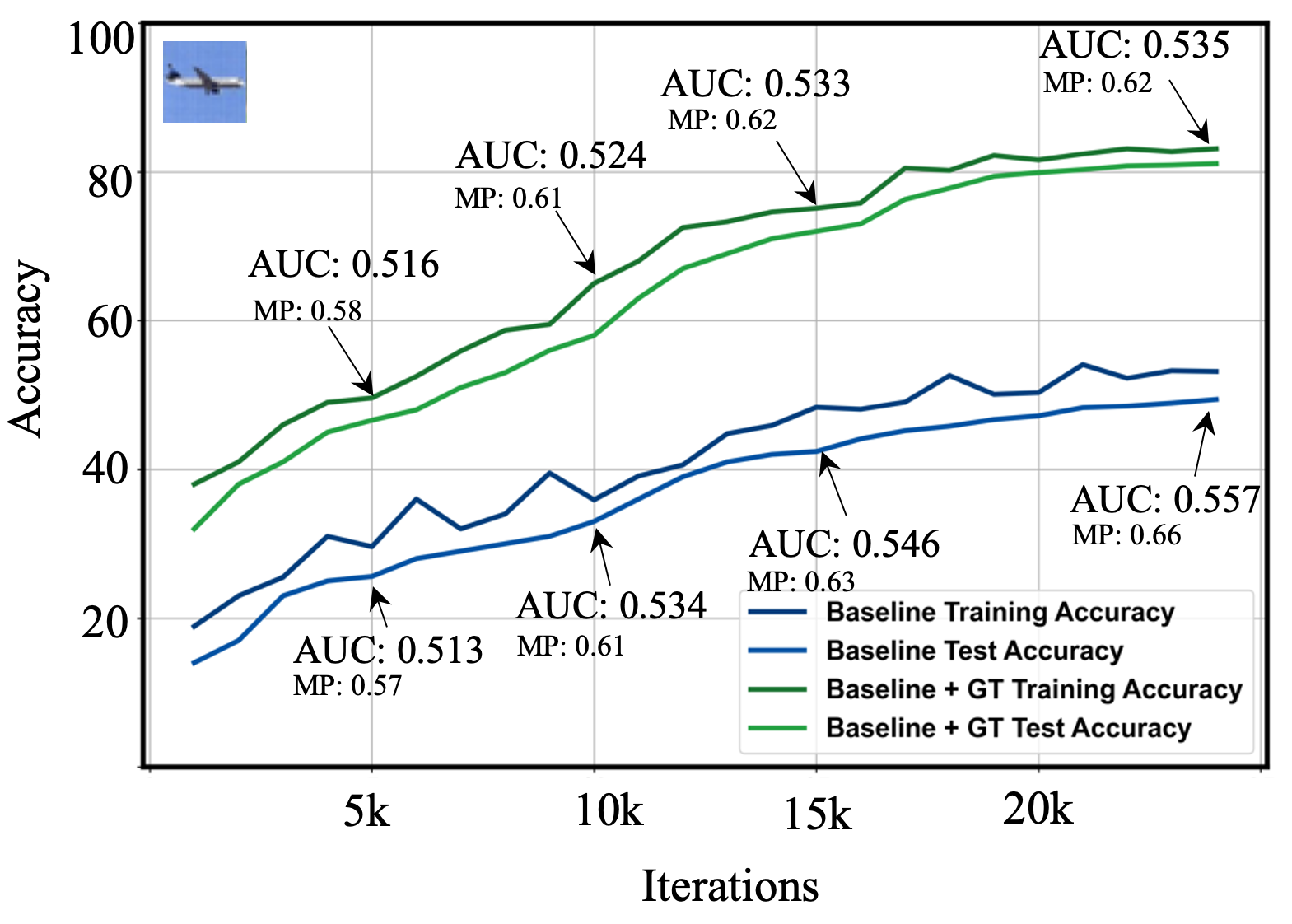}
        \caption{Private learning ($\epsilon=8$)}
        \label{fig:tracking_auc_a}
       \end{subfigure}
       \hfill
       \begin{subfigure}{0.48\linewidth}
        \includegraphics[width=\textwidth]{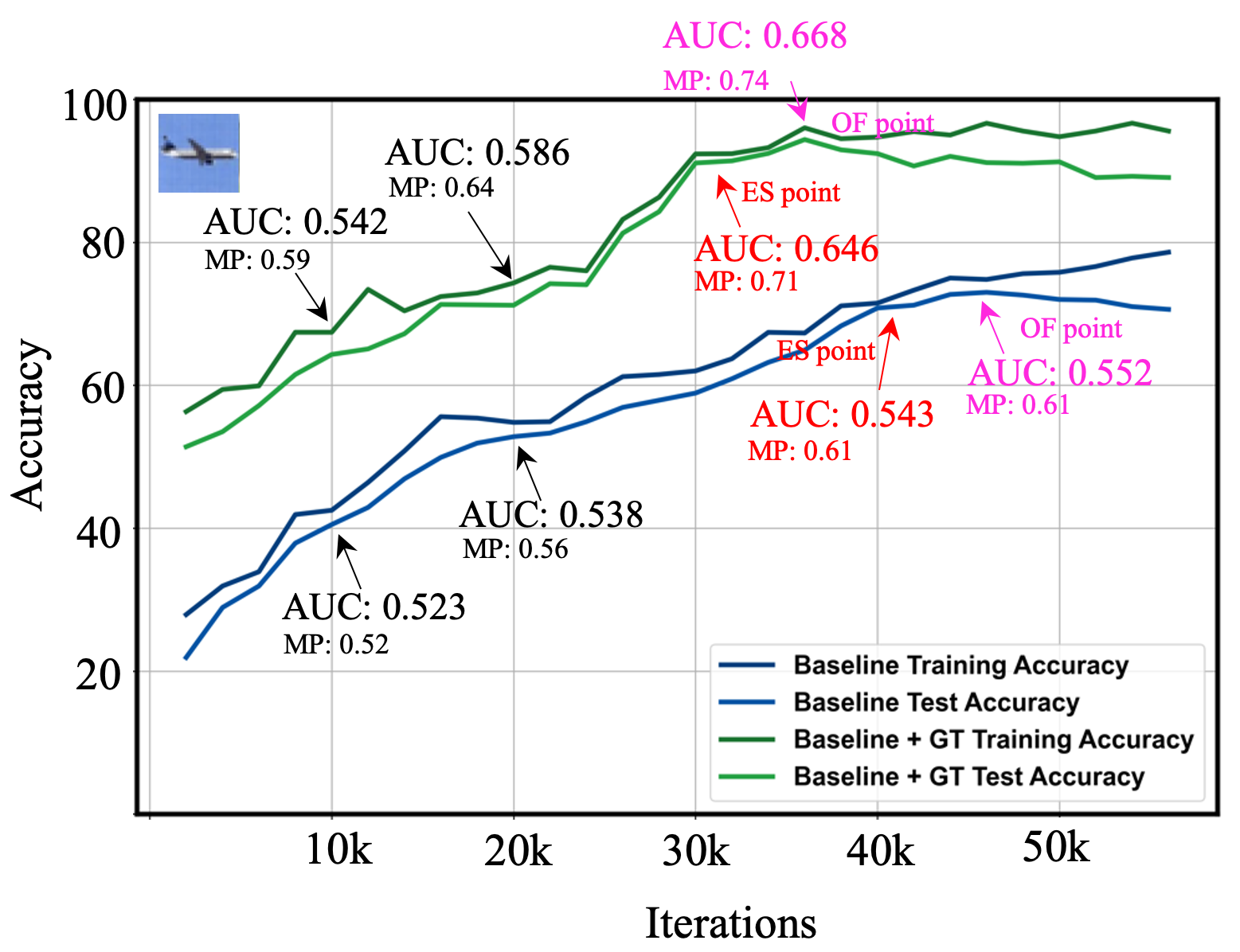}
        \caption{Non-private learning}
        \label{fig:tracking_auc_b}
    \end{subfigure}
    \hfill

    \caption{The training and test accuracy for four training processes are depicted for the CIFAR-10 dataset. We track MIA AUC (AUC) and membership probability (MP) for a specific training sample (airplane image shown in the upper left corner of each plot) for a baseline model and after adding generalization techniques (GT). Early stopping (ES) and overfitting (OF) points in the non-private training process are indicated. }
    \label{fig:tracking_auc}
\end{figure}

In our work, despite using early stopping techniques to prevent overfitting, we observe evidence of this memorization. We track MIA AUC throughout the training process, both for private and non-private models, across baseline models and after incorporating all generalization techniques. In Figure~\ref{fig:tracking_auc}, we track a specific training sample with a high membership (inference) probability (MP) throughout the training processes. We observe that this sample is memorized well before the stopping point indicated by the early stopping methods. This finding suggests that, even with generalization techniques in place, certain samples are memorized early in the training phase, highlighting that generalization techniques may leave specific samples more susceptible to privacy risks.


\subsubsection{The Comparison of DP-SGD and DP-SAT}

Table~\ref{dp_sat_acc} compares DP-SGD and DP-SAT for different privacy budgets when both methods use the generalization techniques of~\cite{de2022unlocking}. DP-SAT exhibits superior classification accuracy consistently compared to DP-SGD in all scenarios. It shows that the identification of flat minima becomes notably advantageous. 
\begin{table}
\centering
\caption{The accuracy and standard deviation (in gray) of models trained on CIFAR-10 and CIFAR-100 using DP-SGD and DP-SAT.}
\label{dp_sat_acc}
\begin{tabular}{|c|c|c|c|} 
\hline
Datasets                                                                                              & \begin{tabular}[c]{@{}c@{}}Privacy Budget $\epsilon$ \\$(\delta = 10^{-5})$\end{tabular} & DP-SGD                                          & DP-SAT                                           \\ 
\hline
\multirow{4}{*}{CIFAR-10}                                                                             & $\epsilon = 1$                                                                  & 45.13 \textcolor[rgb]{0.502,0.502,0.502}{(0.1)} & 45.94 \textcolor[rgb]{0.502,0.502,0.502}{(0.1)}  \\ 
\cline{2-4}
 & $\epsilon = 2$                                                                  & 60.23 \textcolor[rgb]{0.502,0.502,0.502}{(0.2)} & 62.59 \textcolor[rgb]{0.502,0.502,0.502}{(0.2)}  \\ 
\cline{2-4}
 & $\epsilon = 4$                                                                  & 69.74 \textcolor[rgb]{0.502,0.502,0.502}{(0.4)} & 72.1 \textcolor[rgb]{0.502,0.502,0.502}{(0.3)}   \\ 
\cline{2-4}
 & $\epsilon = 8$                                                                  & 79.06 \textcolor[rgb]{0.502,0.502,0.502}{(0.7)} & 81.11 \textcolor[rgb]{0.502,0.502,0.502}{(0.3)}  \\ 
\hline
\multirow{4}{*}{\begin{tabular}[c]{@{}c@{}}CIFAR-100\\\end{tabular}} & $\epsilon = 1$                                                                  & 69.98 \textcolor[rgb]{0.502,0.502,0.502}{(0.3)} & 71.71 ~\textcolor[rgb]{0.502,0.502,0.502}{(0.2)}  \\ 
\cline{2-4}
 & $\epsilon = 2$                                                                  & 76.17~\textcolor[rgb]{0.502,0.502,0.502}{(0.6)} & 77.8 \textcolor[rgb]{0.502,0.502,0.502}{(0.4)}    \\ 
\cline{2-4}
& $\epsilon = 4$                                                                  & 79.28 \textcolor[rgb]{0.502,0.502,0.502}{(0.4)}  & 81.83 ~\textcolor[rgb]{0.502,0.502,0.502}{(0.3)}  \\ 
\cline{2-4}
& $\epsilon = 8$                                                                  & 83.1 ~\textcolor[rgb]{0.502,0.502,0.502}{(0.3)} & 83.8 ~\textcolor[rgb]{0.502,0.502,0.502}{(0.2)}   \\
\hline
\end{tabular}
\end{table}

\subsection{Fairness-Utility Trade-off}
\begin{table}
\caption{Measured bias for models with $(8, 10^{-5})$-DP and without DP trained on CIFAR-10S and CIFAR-100S.}
\centering
\label{cifar_10s_bias}
\begin{tabular}{|l|clcl|clcl|}
\hline
\multirow{3}{*}{} & \multicolumn{4}{c|}{CIFAR-10S}                                                                   & \multicolumn{4}{c|}{CIFAR-100S}                                                                  \\ \cline{2-9} 
                  & \multicolumn{2}{l|}{\multirow{2}{*}{With DP}} & \multicolumn{2}{l|}{\multirow{2}{*}{Wihtout DP}} & \multicolumn{2}{l|}{\multirow{2}{*}{With DP}} & \multicolumn{2}{l|}{\multirow{2}{*}{Without DP}} \\
                  & \multicolumn{2}{l|}{}                         & \multicolumn{2}{l|}{}                            & \multicolumn{2}{l|}{}                         & \multicolumn{2}{l|}{}                            \\ \hline
Baseline          & \multicolumn{2}{c|}{0.04}                     & \multicolumn{2}{c|}{0.01}                        & \multicolumn{2}{c|}{0.10}                     & \multicolumn{2}{c|}{0.11}                        \\ \hline
+ GN              & \multicolumn{2}{c|}{0.06}                     & \multicolumn{2}{c|}{0.03}                        & \multicolumn{2}{c|}{0.12}                     & \multicolumn{2}{c|}{0.12}                        \\ \hline
+ OBS              & \multicolumn{2}{c|}{0.07}                     & \multicolumn{2}{c|}{0.03}                        & \multicolumn{2}{c|}{0.15}                     & \multicolumn{2}{c|}{0.13}                        \\ \hline
+ WS              & \multicolumn{2}{c|}{0.09}                     & \multicolumn{2}{c|}{0.05}                        & \multicolumn{2}{c|}{0.19}                     & \multicolumn{2}{c|}{0.17}                        \\ \hline
+ AM              & \multicolumn{2}{c|}{0.11}                     & \multicolumn{2}{c|}{0.07}                        & \multicolumn{2}{c|}{0.23}                     & \multicolumn{2}{c|}{0.19}                        \\ \hline
+ PA              & \multicolumn{2}{c|}{0.14}                     & \multicolumn{2}{c|}{0.08}                        & \multicolumn{2}{c|}{0.26}                     & \multicolumn{2}{c|}{0.21}                        \\ \hline
+ SAT             & \multicolumn{2}{c|}{0.15}                     & \multicolumn{2}{c|}{0.09}                        & \multicolumn{2}{c|}{0.27}                     & \multicolumn{2}{c|}{0.21}                        \\ \hline
\end{tabular}
\end{table}
\begin{table*}
\centering
\caption{Measured accuracy (acc), MIA AUC (AUC), bias, and HS in models trained on CIFAR-10S and CIFAR-100S in both private and non-private settings.}
\label{cifar10s100s_acc_att_bias}
\begin{tabular}{|c|c|c|c|l|c|c|c|l|} 
\hline
\multirow{2}{*}{} & \multicolumn{4}{c|}{$(8, 10^{-5})$-DP}                       & \multicolumn{4}{c|}{Without DP}              \\ 
\cline{2-9}
                  & acc (C\textbar{}G)   & AUC & bias  & HS & acc  (C\textbar{}G) & AUC & bias & HS  \\ 
\hline
CIFAR-10S         & 65.41\textbar{}65.25 & 0.82     & 0.15  & 0.525   & 86.8\textbar{}85.1  & 0.85     & 0.09 & 0.532    \\ 
\hline
CIFAR-100S        & 37.2\textbar{}37.3   & ~0.68    & ~0.27 & 0.618   & 72.1\textbar{}72.3  & ~0.71    & 0.21 & 0.566    \\
\hline
\end{tabular}
\end{table*}
This section measures model bias, according to Eq.~\ref{eq:bias_metric1}, while applying generalization techniques in private (i.e., $(8, 10^{-5})$-DP) and non-private models trained on biased datasets CIFAR-10S and CIFAR-100S. We cumulatively apply each generalization technique into the SGD training process, replace SGD with SAT as the final generalization technique in our sequence, and compute the model bias after each integration. 

Although Table~\ref{cifarS-privacy} demonstrates that the generalization techniques improve accuracy for models trained on biased datasets CIFAR-10S and CIFAR-100S, Table~\ref{cifar_10s_bias} shows this improvement is accompanied by a significant increase in the models' bias. Specifically, in the non-private model trained on CIFAR-10S, the bias metric has escalated by a factor of 9, from 0.01 to 0.09. For the private model trained on CIFAR-10S, the bias metric has escalated by a factor of 3.75, from 0.04 to 0.15. This trend of increasing model bias is also apparent in CIFAR-100S, with a bias rise by a factor of 2.7 in the non-private model, from 0.11 to 0.27, and by a factor of 2.6 in the private model, from 0.10 to 0.26. As it is shown in Table~\ref{cifarS-privacy}, for CIFAR-10S, AM and WS amplify bias more than other techniques in the non-private model, while PA is the main contributor to increasing bias in the private context. For CIFAR-100S, both WS and AM play crucial roles in adjusting bias in non-private and private models.
\begin{figure*}
    \centering
 \begin{subfigure}{0.3\linewidth}
        \includegraphics[width=\textwidth]{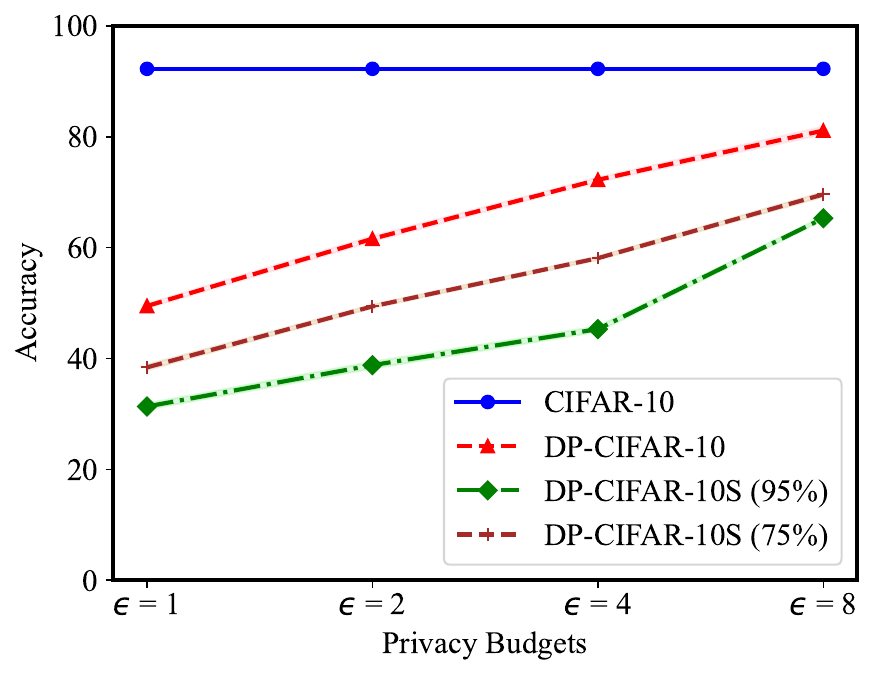}
        \caption{}
        \label{fig:accuracy_cifar10}
\end{subfigure}  
\hfill
 \begin{subfigure}{0.3\linewidth}
        \includegraphics[width=\textwidth]{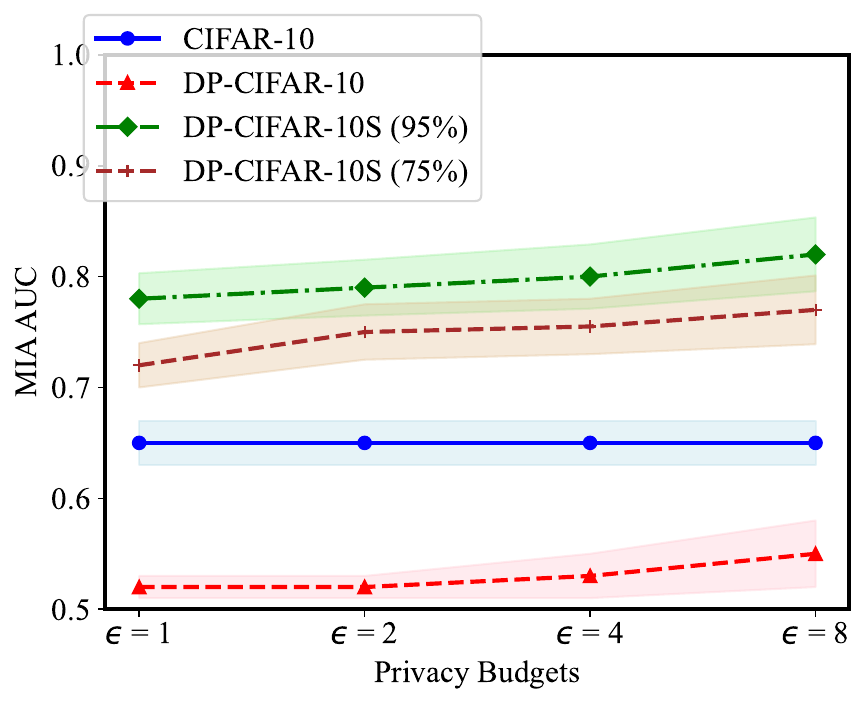}
        \caption{}
        \label{fig:attack_auc_cifar10}
\end{subfigure}
\hfill
 \begin{subfigure}{0.3\linewidth}
     \includegraphics[width=\textwidth]{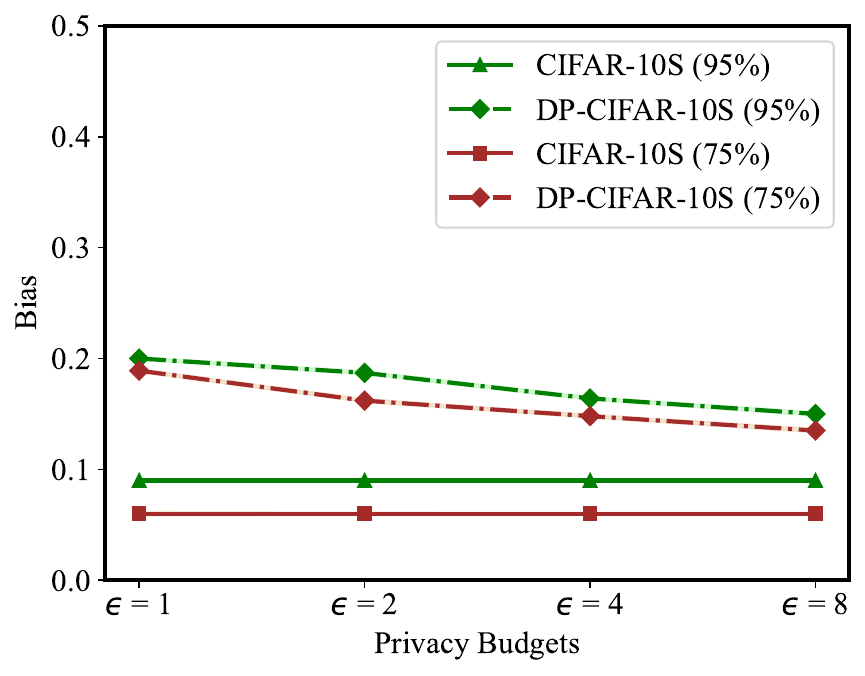}
     \caption{}
        \label{fig:bias_cifar10}
\end{subfigure}
\hfill
 
    \caption{Examined the impact of different privacy budgets ($\epsilon$ = 1, 2, 4, 8) and data bias (95\%, 75\%) on (a) accuracy, (b) MIA AUC, (c) bias. DP-CIFAR-10 and DP-CIFAR-10S are used to denote when DP is applied.}
    \label{fig:acc_auc_bias}
\end{figure*}
\begin{figure}

    \includegraphics[width=0.8\linewidth]{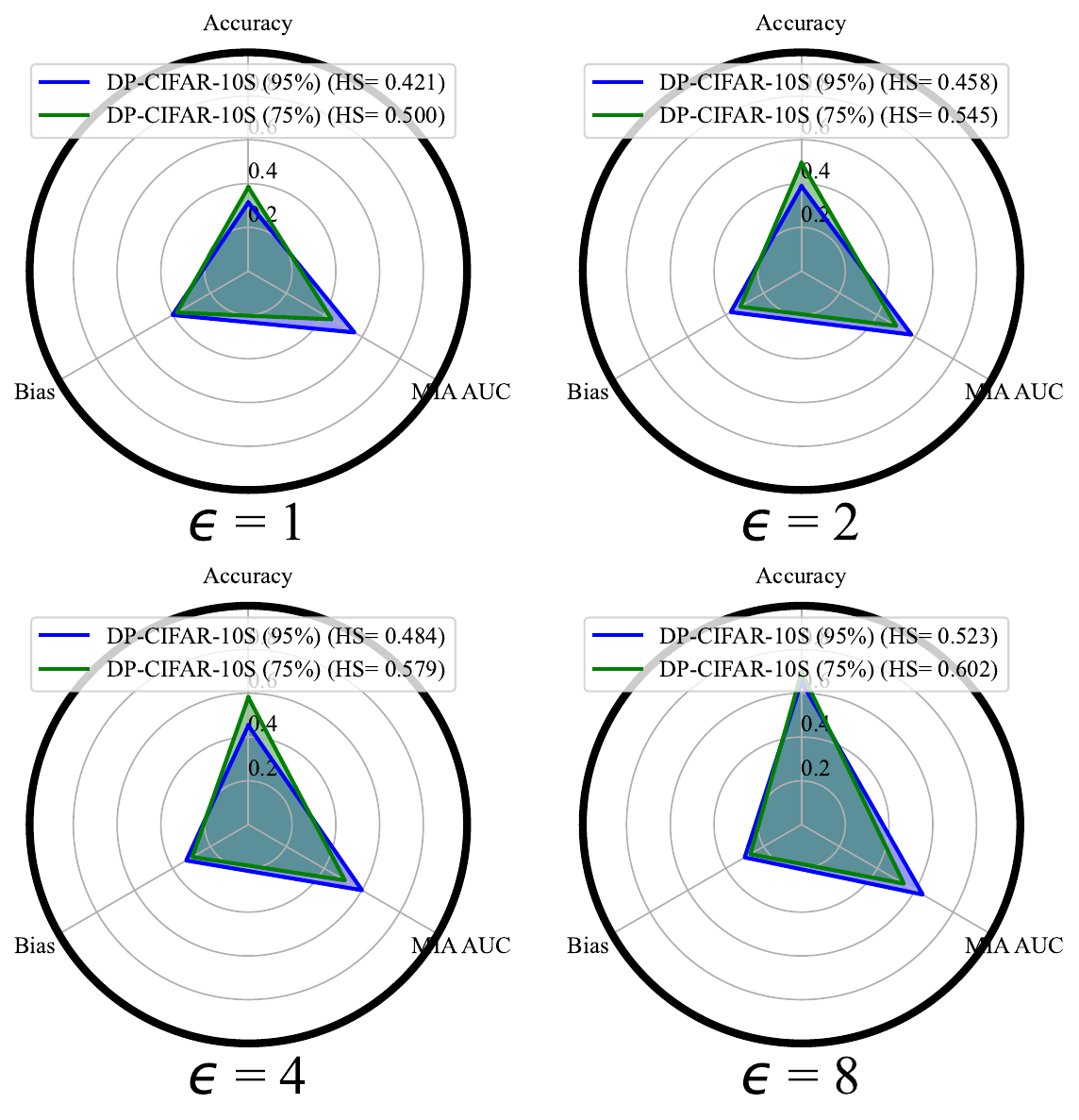}
    \caption{Examined the impact of different privacy budgets ($\epsilon$ = 1, 2, 4, 8) and data bias (95\%, 75\%) on HS.}
    \label{fig:overall_abe_cifar10}

\end{figure}
\subsection{Privacy and Fairness-Utility Trade-offs}
This section analyzes the relationship between accuracy, privacy, and fairness in models trained with and without DP on biased datasets, CIFAR-10S and CIFAR-100S (addressing \textbf{Q2}). We explore how this relationship evolves across different privacy budgets $(\epsilon = 1,2,4,8)$ and dataset bias levels, specifically at 95\% and 75\%. Additionally, we investigate how the HS of a model fluctuates under varying conditions of privacy settings and data bias. By examining these variables, we aim to provide a comprehensive understanding of the trade-offs involved in differentially private learning under various scenarios.

As demonstrated in Table~\ref{cifar10s100s_acc_att_bias}, the model bias (computed by Eq.~\ref{eq:bias_metric1}) for the model trained with DP on CIFAR-10S is 0.15, 0.06 higher than the non-private model. Even though the private model reduces MIA AUC by a marginal 0.03, it significantly drops the overall accuracy by approximately 21\%. This significant drop in accuracy indicates a substantial trade-off when implementing DP, highlighting the challenge of maintaining utility while ensuring privacy. Similarly, examining the model trained on CIFAR-100S with DP, the bias metric increased by 0.06 compared to the scenario without DP. This increase in model bias, although indicative of potential fairness issues, accompanies a 10\% decrease in MIA AUC, suggesting an improvement in privacy. These findings underscore the intricate balance that must be navigated between improving privacy and maintaining model accuracy, particularly in biased datasets. For CIFAR-10S, the comparison of HSs exhibits how DP, even with a relatively lenient privacy budget of $\epsilon = 8$, disrupts the delicate balance between accuracy, MIA AUC, and bias. For CIFAR-100S, the introduction of DP significantly improves privacy and consequently helps the balance between accuracy, MIA AUC, and bias.

\begin{figure*}
    \centering
      \begin{subfigure}{0.3\linewidth}
        \includegraphics[width=\textwidth]{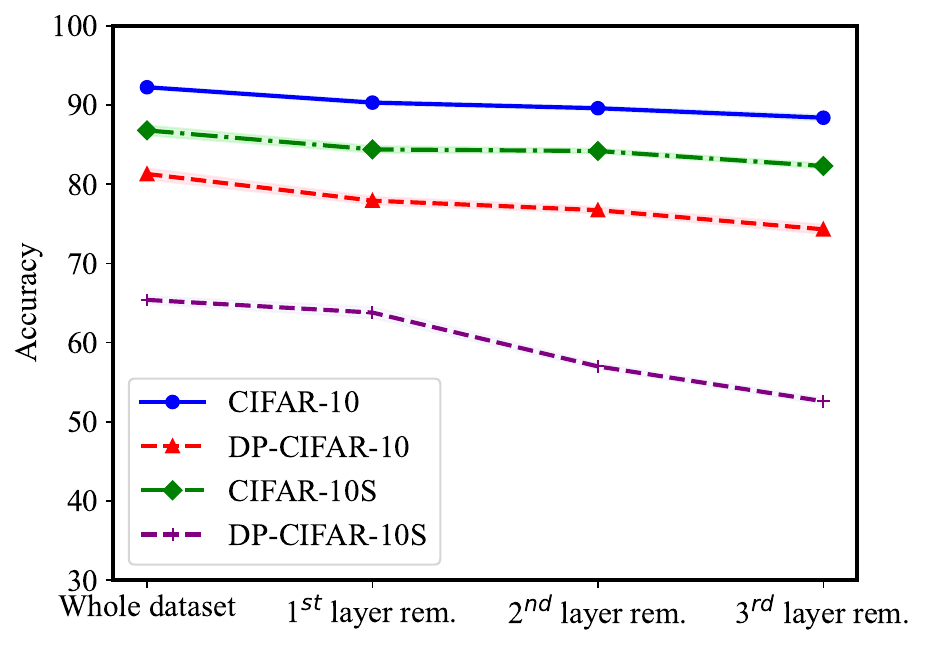}
        \caption{}
        \label{fig:accuracy_onion_effect_cifar10}
       \end{subfigure}
       \hfill
       \begin{subfigure}{0.3\linewidth}
        \includegraphics[width=\textwidth]{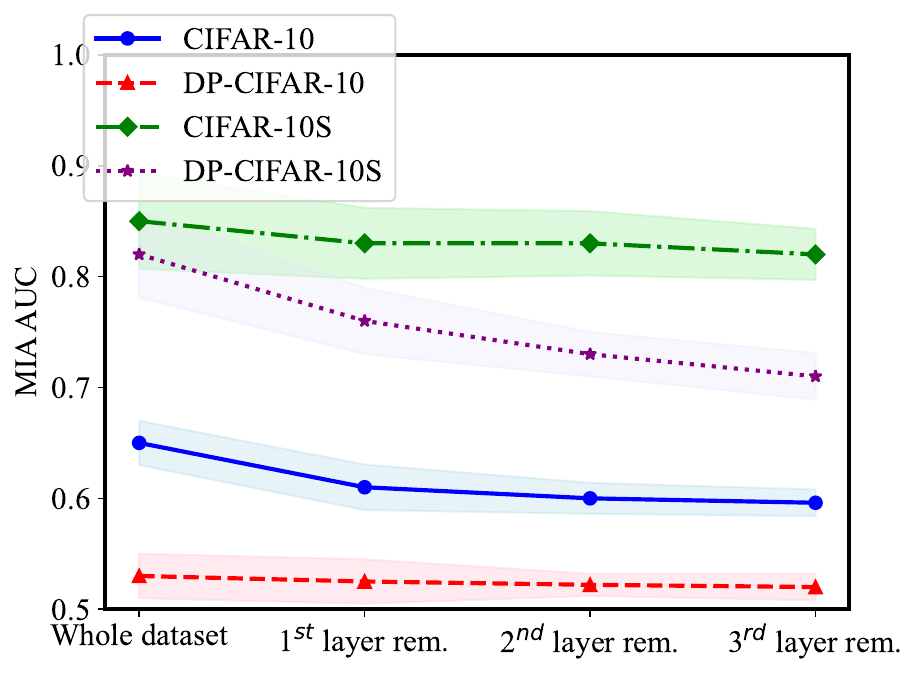}
        \caption{}
        \label{fig:attack_acc_onion_effect_cifar10}
    \end{subfigure}
    \hfill
      \begin{subfigure}{0.3\linewidth}
        \includegraphics[width=\textwidth]{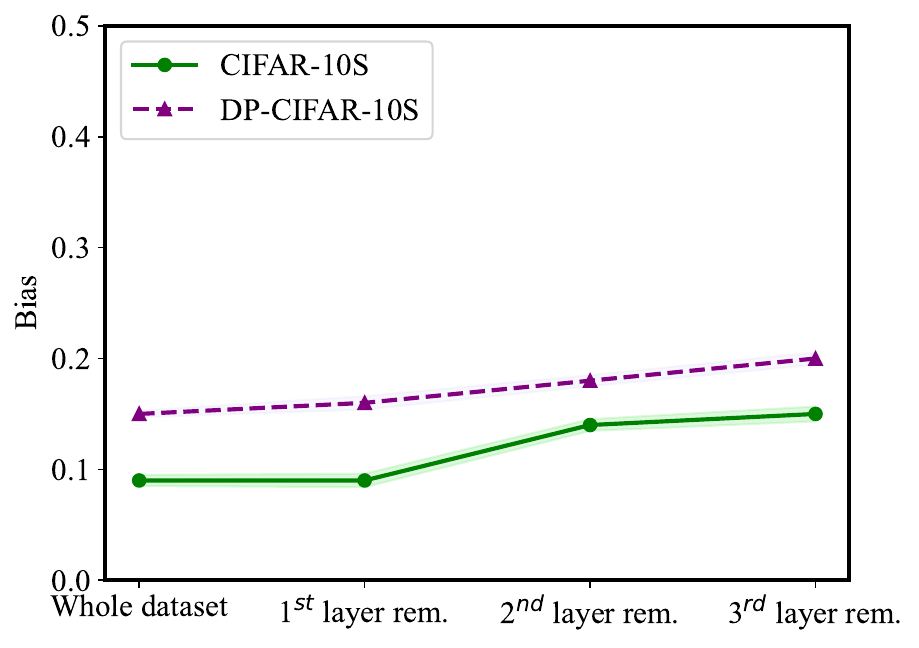}
        \caption{}
        \label{fig:bias_onion_effect_cifar10}
    \end{subfigure}
    \hfill
       
    \caption{The impact of removing outliers on model performance (a) accuracy, (b) MIA AUC, and (c) bias trained with and without DP on CIFAR-10 and CIFAR10S. DP-CIFAR-10 and DP-CIFAR-10S are used to denote when DP is applied.}
    \label{fig:onion_effectcifar10}
\end{figure*}
\begin{figure}

        \includegraphics[width=\linewidth]{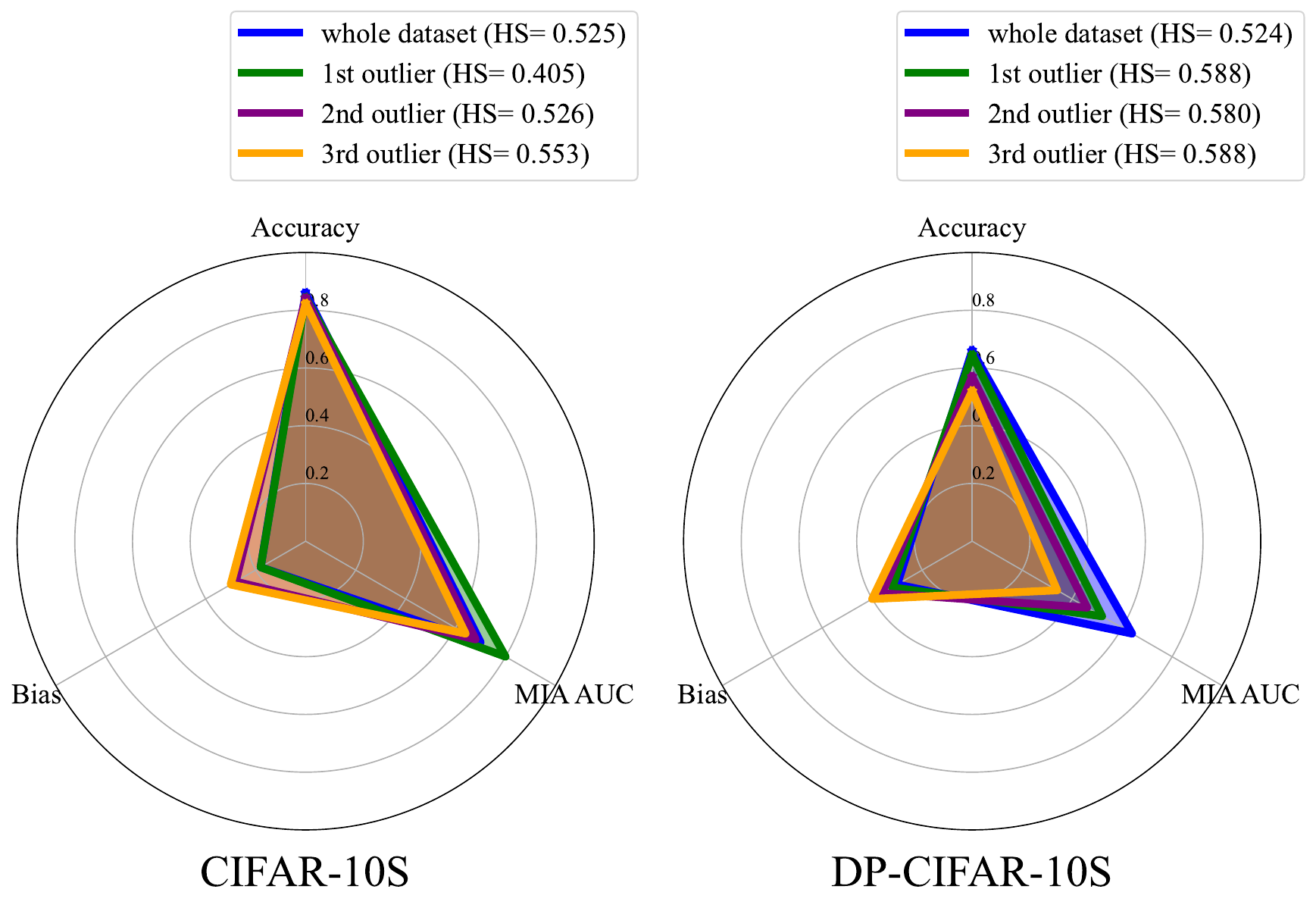}
        \caption{HS evaluation on CIFAR-10S and DP-CIFAR-10S datasets after exclusion of three outlier layers.}
        \label{fig:abe_onion_effect_cifar10}
   
\end{figure}
Figure~\ref{fig:accuracy_cifar10} illustrates that an increase in the training data bias (from 75\% to 95\%) corresponds to a decrease in accuracy for all $\epsilon$ values. This can be attributed to the reduced representativeness and increased skewness of CIFAR10S (95\%). When the data is heavily skewed towards a particular outcome, the added noise disproportionately impacts the less represented data, making accurate predictions more challenging. This decrease in accuracy highlights the adverse effects of training data bias on model performance, further complicating the privacy-utility trade-off. 

Figure~\ref{fig:attack_auc_cifar10} demonstrates that higher bias in the training data increases MIA AUC across all $\epsilon$ values. This suggests bias boosts privacy risks by highlighting data patterns that may reveal an individual's membership, regardless of the privacy budget value. The increased MIA AUC with higher bias indicates that data bias not only affects utility but also significantly heightens privacy risks, complicating efforts to protect sensitive information. Figure~\ref{fig:attack_auc_cifar10} also reveals that biased datasets, even at lower privacy budgets, can still be highly vulnerable to MIAs. Interestingly, the models trained with DP on the biased CIFAR-10S—regardless of whether the bias level is set at 95\% or 75\% — demonstrate more vulnerability to MIAs than the model trained without DP on the unbiased CIFAR-10. This suggests that the effectiveness of DP in protecting individual data points is highly compromised when the training data is biased, highlighting a significant limitation in current DP methods.

Figure~\ref{fig:bias_cifar10} depicts that a boost in the training data bias amplifies the model bias for all $\epsilon$ values. Furthermore, despite beginning at a high value, the bias in model outcome shows a slight reduction as the epsilon rises. This reduction, albeit small, indicates that increasing the privacy budget can slightly mitigate the exacerbation of bias, though it does not fully resolve the fairness issues introduced by biased training data.

Figure~\ref{fig:overall_abe_cifar10} demonstrates that increasing privacy budgets and reducing the training data bias result in higher HS, offering a better balance between accuracy, MIA AUC, and bias. These results suggest that careful adjustment of privacy budgets and efforts to reduce (training) data bias are essential for achieving a more balanced model performance. 

Similarly, models trained on CIFAR-100 and CIFAR-100S show consistent patterns and findings; see the Supplementary material Figure~\ref{fig:acc_auc_bias_cifar100} for clarity.

\subsection{Onion Effect Impact}
This section explores the Onion Effect~\cite{carlini2022privacy}, a phenomenon where peeling away a layer of outlier samples that are most vulnerable to a privacy attack unveils a subsequent layer of samples newly vulnerable to the same attack. Our investigation extends and differentiates from the work of~\cite{carlini2022privacy} by conducting a broader analysis. This includes examining models trained without DP not only on the unbiased CIFAR-10 but also those trained without DP on the unbiased CIFAR-100 and models trained with and without DP on the biased CIFAR-10S and CIFAR-100S. We evaluate not only MIA AUC but also the models' accuracy when relevant, bias, and HS. By considering these diverse datasets and training scenarios, our analysis provides a comprehensive view of how the Onion Effect impacts various aspects of model performance and privacy.

Additionally, we intensify the scrutiny of the Onion Effect by removing three layers of outliers. In our experiment, outliers are discerned through their privacy risk scores, computed by the MLP attack algorithm~\cite{tfplib}. The privacy risk score of an input sample for a target model reflects the posterior probability that the sample is part of the model's training data, inferred from the model’s behavior in response to that sample~\cite{song2021systematic}. This methodical approach ensures that the outliers identified are those that the MIA attacks classify as members with the highest confidence scores.

Our experimental process includes three rounds of data removal, each eliminating 5,000 samples, equating to a 10\% reduction in the first phase, 11.1\% in the second, and 12.5\% in the third. After each removal, we retrain a new model to evaluate model accuracy, MIA AUC, bias (when relevant), and HS. This iterative removal and retraining process allows us to observe the cumulative effects of outlier removal on the model’s performance and privacy metrics.

Figure~\ref{fig:accuracy_onion_effect_cifar10} shows removing outlier layers decreases accuracy, suggesting such removals deprive the model of essential information for more precise predictions. The decline in accuracy underscores the critical role that even the most vulnerable samples play in the overall predictive capability of the model.

Figure~\ref{fig:attack_acc_onion_effect_cifar10} illustrates that MIA AUC stays the same or goes slightly down in private and non-private models. This describes the Onion Effect, continuously exposing vulnerable data points as layers of outliers are peeled away. The persistence of high MIA AUC values indicates that removing outliers does not mitigate the inherent vulnerability of the remaining data, confirming the findings of Carlini et al.~\cite{carlini2022privacy} even though their attack type differs from ours. A reason for this effect lies in counterfactual influence~\cite{carlini2022privacy}: the model’s behavior toward a target data point can be significantly shaped by the presence of nearby samples in feature space. When the most influential outliers are removed, previously stable inliers near the boundary of the data distribution gain higher counterfactual influence, effectively becoming the new outliers. This shift increases the membership inference advantage for these remaining points, as their prediction behaviors now exhibit heightened sensitivity to changes in the training set composition, making them more vulnerable to privacy attacks.

Figure~\ref{fig:bias_onion_effect_cifar10} indicates removing outliers may increase bias, likely because discarding outlier layers lowers accuracy for underrepresented groups, exacerbating bias in the model's predictions. This increase in bias reveals a potential trade-off between privacy and fairness, where efforts to protect privacy might inadvertently harm model equity.

Finally, Figure~\ref{fig:abe_onion_effect_cifar10} shows that outlier removals can be useful in balancing accuracy, privacy, and fairness, particularly in private contexts. The improvement in HS values after outlier removal suggests that such interventions may provide a viable solution for achieving a balanced optimization of these key metrics. 

See Supplementary material Figure~\ref{fig:onion_effectcifar100} for CIFAR-100 and CIFAR-100S results, supporting the above-discussed conclusions.

\section{Real-World Setting: Validation with CelebA Attributes}

To validate our findings from synthetic datasets, we use the CelebA dataset for a real-world evaluation of facial attribute recognition. Following~\cite{wang2020towards}, our evaluation uses mean average precision (mAP) to measure classification performance, adjusted for gender balance with a weighted mAP metric: for an attribute present in the images of \(N_m\) men and \(N_w\) women, we weight positive images of men by \(\frac{(N_m + N_w)}{(2N_m)}\) and positive images of women by \(\frac{(N_m + N_w)}{(2N_w)}\). This adjustment ensures equal consideration of positive samples from men and women. Additionally, we apply the bias metric, as described in Eq.~\ref{eq:bias_metric2}, to assess how the model’s predictions might amplify existing gender imbalances. 
This approach enables us to assess the model’s ability to handle real-world biases and validate the trends observed in controlled, synthetic settings.

Table~\ref{tab:celebA} compares performance metrics such as mAP, bias, and MIA AUC across varying skewness levels for non-private and private (i.e., $(8, 10^{-5})$-DP) models trained on the CelebA attribute dataset. Integrating generalization techniques increases the mAP of both the non-private \textit{Baseline} model and the private \textit{Baseline + $(8, 10^{-5})$-DP}, aligning with the trend observed in Table~\ref{cifarS-privacy}. However, this integration also amplifies bias in both the non-private \textit{Baseline} and private \textit{Baseline + $(8, 10^{-5})$-DP}, consistent with the trend shown in Table~\ref{cifar_10s_bias}. Furthermore, observing higher MIA AUC values for attributes with greater skewness ($\geq 0.8$) supports the conclusion from Figure~\ref{fig:bias_cifar10} and Figure~\ref{fig:bias_cifar100} that increased training data bias makes models more susceptible to MIAs.

\begin{figure}
    \centering
    \includegraphics[width=1\linewidth]{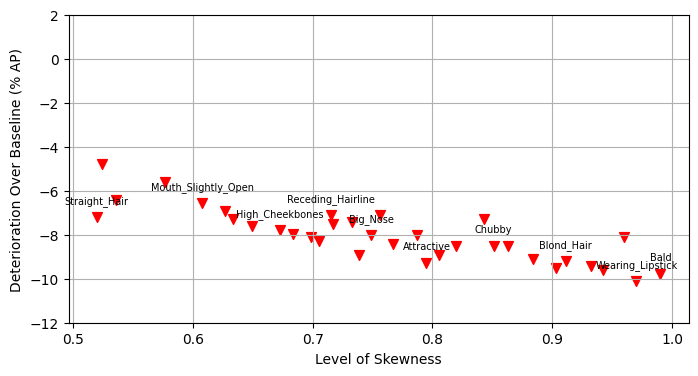}
    \caption{Per-attribute average precision (AP) deterioration of the DP-enabled with generalization techniques model over the Baseline model on the CelebA validation set, as a function of the level of gender imbalance in the attribute. Attributes with high skewness (such as “bald”) suffer most significantly.}
    \label{fig:celebA}
\end{figure}

Figure ~\ref{fig:celebA} illustrates the relationship between the skewness level of various attributes in the CelebA dataset and their corresponding improvement over the baseline model, measured in terms of Average Precision (AP). The overall trend demonstrates a consistent decline in AP as the skewness level increases. Specifically, for attributes with a skewness level greater than or equal to 0.8, the AP decreases by -8.96, suggesting that these attributes are more adversely affected by the model's bias. This score, for attributes with a skewness level less than or equal to 0.8, is -7.43, indicating that attributes with lower skewness experience less negative impact. 

\begin{table}
\small
\setlength{\tabcolsep}{2pt}
\centering
\begin{tabular}{|l|c|c|c|c|}
\hline
Model & mAP & bias & \multicolumn{2}{c|}{Average AUC}  \\ \cline{4-5}
& & & 0.5 $\leq$ skew. $\leq$ 1 & skew. $\geq 0.8$ \\ \hline
Baseline & 74.3 & 0.011 & 0.62 & 0.69  \\ \hline
\makecell[l]{Baseline + \\ Generalization techniques} & 77.8 & 0.019 & 0.59 & 0.65  \\ \hline
\makecell[l]{Baseline + \\ $(8, 10^{-5})$-DP} & 54.6 & 0.013 & 0.54 & 0.59  \\ \hline
\makecell[l]{Baseline + \\ Generalization techniques \\ + $(8, 10^{-5})$-DP} & 66.1 & 0.024 & 0.53 & 0.55 \\ \hline
\end{tabular}
\caption{Performance metrics of various models on attribute classification on the CelebA test set, including mAP, bias, and MIA AUC. The Average AUC represents the average MIA AUC across all attributes, specifying different ranges of skewness.}
\label{tab:celebA}
\end{table}

\section{Conclusion}
Our research contributed to the differentially private ML field by exploring the balance between accuracy and privacy when generalization techniques are used. We improved model accuracy by integrating DP-SAT with other methods, such as group normalization, optimal batch size, weight standardization, parameter averaging, and augmentation multiplicity. Specifically, we attained a new accuracy milestone of 81.11\% under $(8, 10^{-5})$-DP on CIFAR-10, surpassing the previous benchmark~\cite{de2022unlocking}. Our analysis demonstrated the superior performance of DP-SAT compared to DP-SGD across various privacy parameters and standard image classification benchmarks. These advancements underscored the potential of our approaches to improve the privacy-utility trade-off, significantly boosting model accuracy in private and non-private learning scenarios.

Furthermore, our investigation into the impact of generalization techniques on model bias showed a trade-off between accuracy and fairness. While these techniques enhanced accuracy, they also increased model bias, particularly in models trained on biased datasets. A key finding of our study is that the generalization techniques may not account for the diverse representations in the data. As these methods optimize for overall accuracy, they may excessively improve predictions for the well-represented groups, thus widening the accuracy gap and increasing bias against underrepresented groups. Another insight of our study is that biased data undermines the fairness of models and increases the privacy risks of MIAs, even in the presence of DP. Biased training data can lead to skewed model outputs, which adversaries may exploit to infer membership information about specific individuals in the dataset. Specifically, we showed that even with early stopping and generalization techniques, certain samples are memorized early in the training process, indicating that these techniques may leave specific samples vulnerable to privacy risks. This highlights a limitation in generalization methods for mitigating privacy concerns. We further expand our experiments to a real-world setting using the CelebA attribute dataset, showing that our results consistently align with real-world attribute imbalances.

Additionally, we introduced the HS metric as a novel tool to evaluate the interplay between accuracy, bias, and privacy, providing a framework for assessing model performance. Our findings highlight that reducing training data bias is instrumental in increasing HS, thus enhancing the balance of privacy, accuracy, and fairness. Our study extended the understanding of the Onion Effect, revealing that removing layers of outliers consistently affected model accuracy, privacy, and bias, further complicating the balance between these aspects in ML models.

\subsection{Limitations and Future Work}
Our ablation study investigated the contributions of generalization techniques when added iteratively in a fixed sequence (GN → OBS → WS → AM → PA), as suggested by De et al.~\cite{de2022unlocking}, with the addition of SAT as the final step. This approach provided insight into the cumulative effects of adding these techniques in this specific order. Additionally, we examined the individual impact of each technique, as detailed in Table~\ref{cifar-privacy-gt}. However, our study did not explore the effects of applying these techniques in alternative sequences or evaluating all possible sub-combinations (e.g., GN + OBS without PA). These questions remain directions for future work. Moreover, the empirical analysis was conducted using specific datasets. Future research could extend these findings to other domains, data types, and model architectures. Despite its utility, the HS metric may simplify the interplay between accuracy, privacy, and fairness. Thus, developing advanced metrics to capture the nuanced dynamics between these aspects is important.

\begin{acks}
The work of Ahmad Hassanpour was supported by the Privacy Matters (PRIMA) project under Grant H2020-MSCA-ITN-2019-860315 and SFI Norwegian Center for Cybersecurity in Critical Sectors (NORCICS) project no. 310105. We extend our gratitude to Professor Staal A. Vinterbo for his insightful comments, and we also thank the anonymous reviewers for their thoughtful feedback and constructive suggestions, which have improved our work.
\end{acks}

\bibliographystyle{ACM-Reference-Format}
\bibliography{main}

\appendix
\clearpage
\section{Supplementary Materials}
\setcounter{page}{1}
\subsection{Additional analysis of privacy- and fairness-utility trade-offs}
\begin{figure*}
    \centering
    \begin{subfigure}{0.3\linewidth}
        \includegraphics[width=\textwidth]{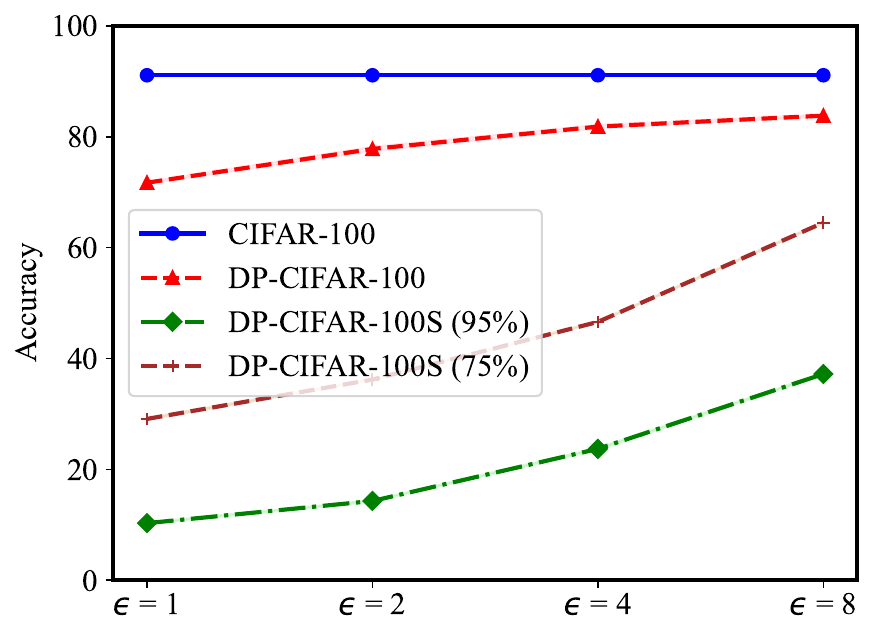}
        \caption{}
        \label{fig:accuracy_cifar100}
    \end{subfigure}
    \hfill
    \begin{subfigure}{0.3\linewidth}
        \includegraphics[width=\textwidth]{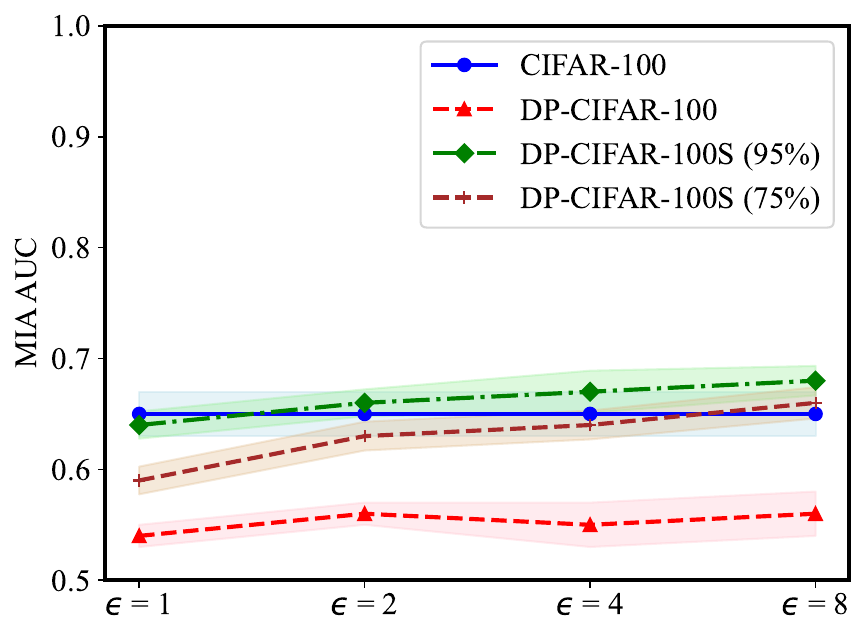}
        \caption{}
        \label{fig:attack_auc_cifar100}
    \end{subfigure}
    \hfill
    \begin{subfigure}{0.3\linewidth}
        \includegraphics[width=\textwidth]{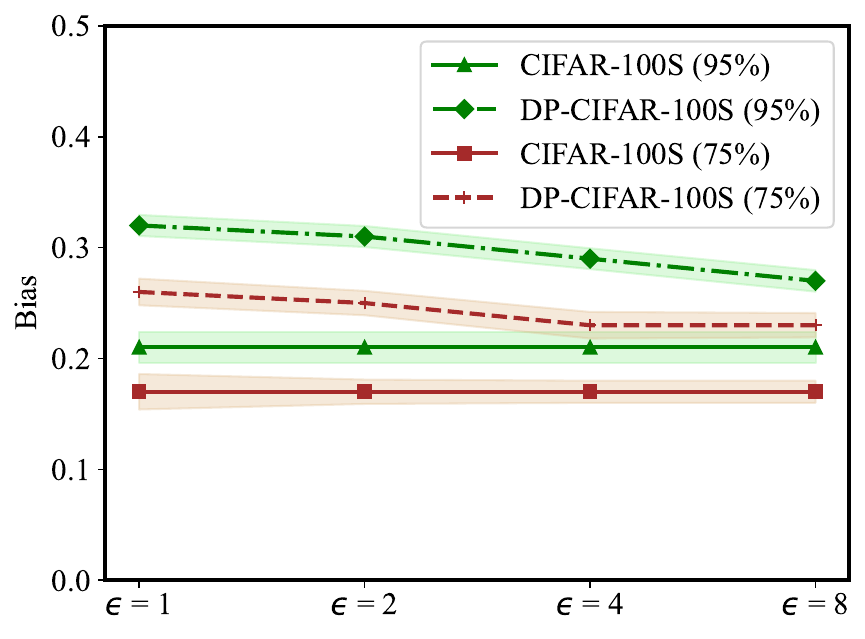}
        \caption{}
        \label{fig:bias_cifar100}
    \end{subfigure}
    \hfill
    \begin{subfigure}{0.4\linewidth}
        \includegraphics[width=\textwidth]{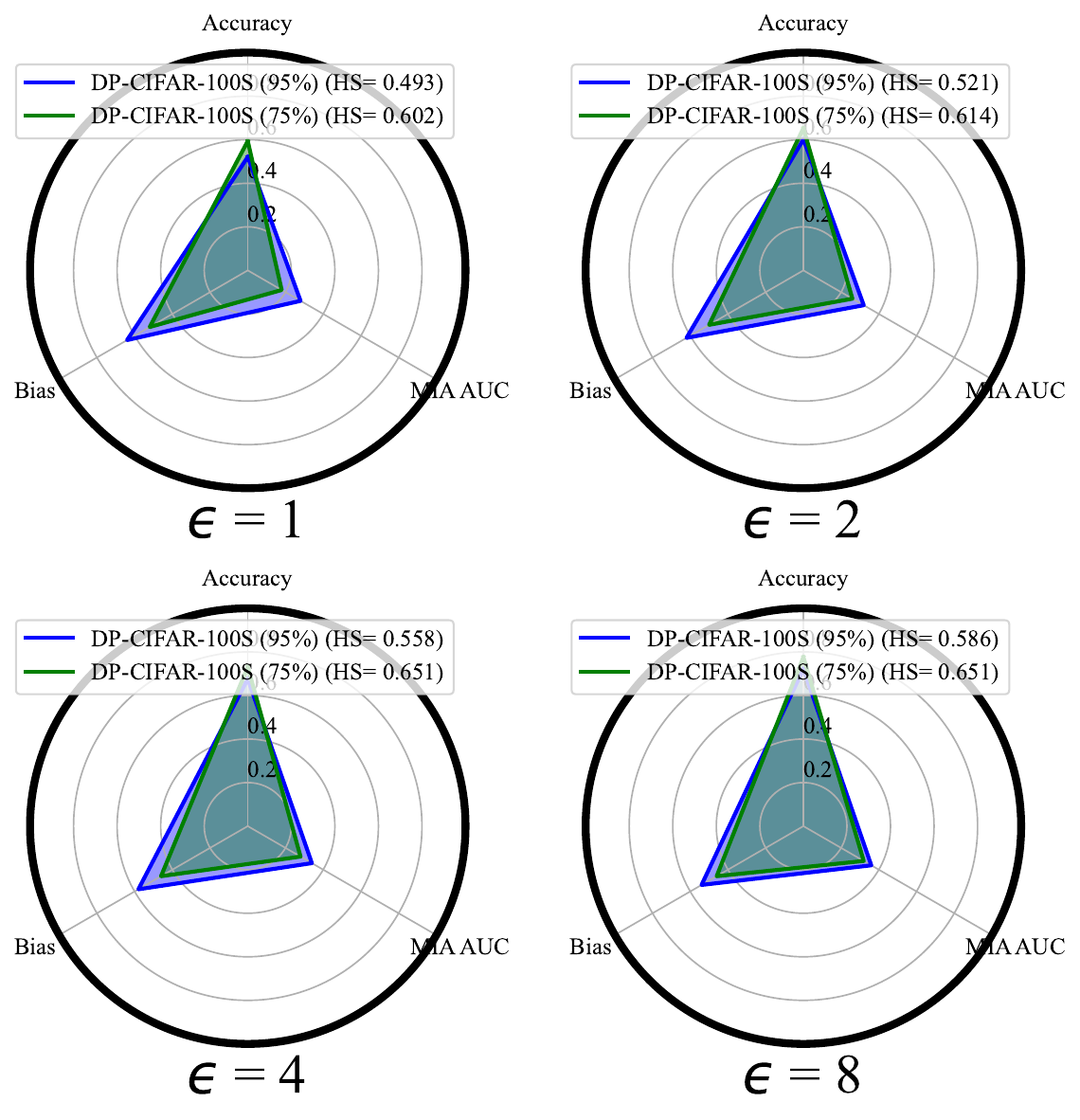}
        \caption{}
        \label{fig:overall_abe_cifar100}
        
    \end{subfigure}
    \caption{Examined the impact of different privacy budgets ($\epsilon$ = 1, 2, 4, 8) and data bias (95\%, 75\%) on (a) accuracy, (b) MIA AUC, (c) bias and (d) HS. DP-CIFAR-100 and DP-CIFAR-100S are used to denote when DP is applied.}
    \label{fig:acc_auc_bias_cifar100}
\end{figure*}

Figure~\ref{fig:acc_auc_bias_cifar100} illustrates the impact of various privacy budgets
$(\epsilon = 1, 2, 4, 8)$ and levels of data bias (95\%, 75\%) on performance metrics (accuracy, MIA AUC, bias, and HS metrics) of models trained with and without
DP on CIFAR-100 and CIFAR-100S datasets. Figure~\ref{fig:accuracy_cifar100} shows increasing training dataset bias (from 75\% to 95\%) lowers accuracy for all $\epsilon$ values due to less representativeness and more skewness in CIFAR10S (95\%). Heavily skewed data affects minority or less represented data, complicating accurate predictions. Figure~\ref{fig:attack_auc_cifar100} finds higher dataset bias raises MIA AUC for all $\epsilon$ values, indicating increased privacy risks through patterns that may signal the individual membership, regardless of privacy budget level. Figure~\ref{fig:bias_cifar100} reveals that rising dataset bias levels amplify model bias across all $\epsilon$ values, and model bias decreases as $\epsilon$ increases. Figure~\ref{fig:overall_abe_cifar100} indicates that higher privacy budgets and reduced training set bias lead to lower HS, enhancing the balance between accuracy, MIA AUC, and bias.
\subsection{Additional analysis of Onion Effect impact}
\begin{figure*}
    \centering
   \begin{subfigure}{0.3\linewidth}
        \includegraphics[width=\textwidth]{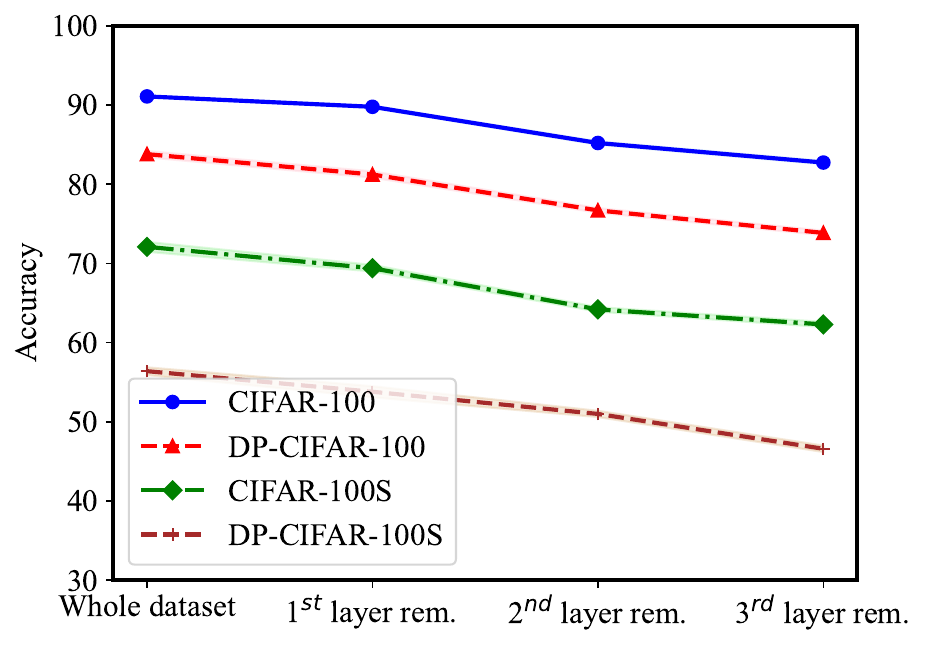}
        \caption{}
        \label{fig:accuracy_onion_effect_cifar100}
    \end{subfigure}
    \hfill
    \begin{subfigure}{0.3\linewidth}
        \includegraphics[width=\textwidth]{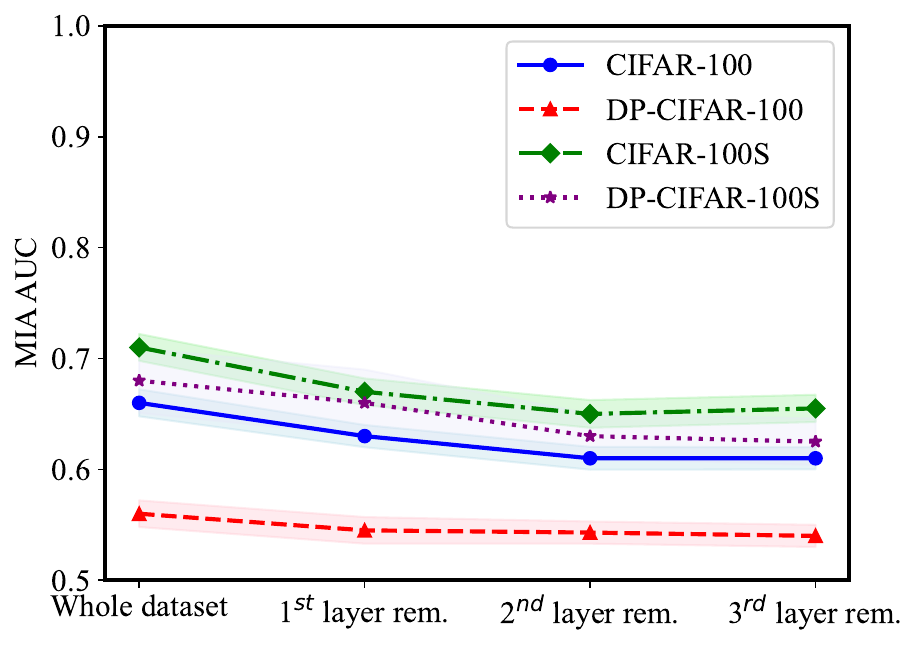}
        \caption{}
        \label{fig:attack_acc_onion_effect_cifar100}
    \end{subfigure}
    \hfill
    \begin{subfigure}{0.3\linewidth}
        \includegraphics[width=\textwidth]{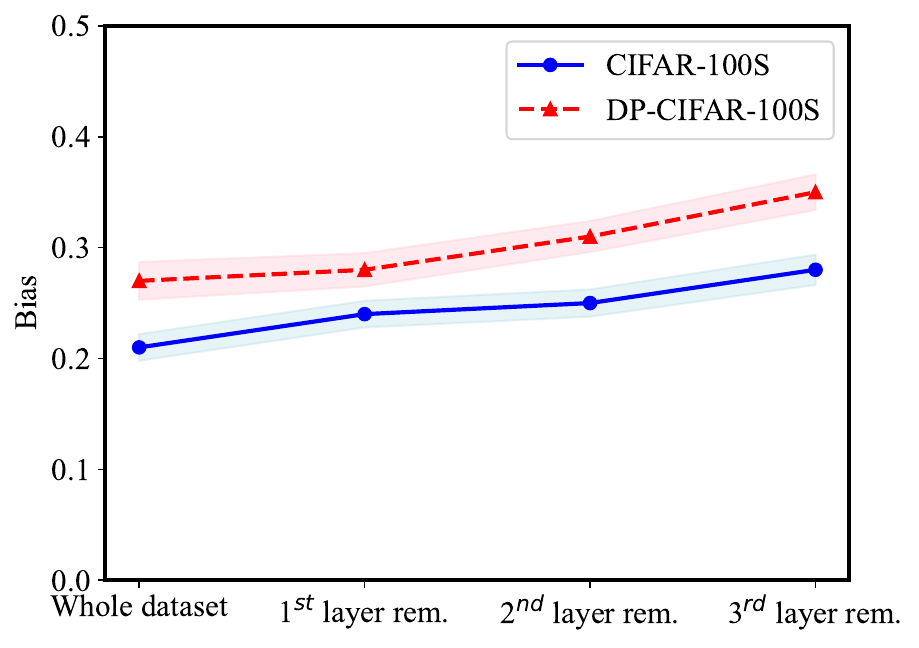}
        \caption{}
        \label{fig:bias_onion_effect_cifar100}
    \end{subfigure}
    \hfill
     \begin{subfigure}{0.4\linewidth}
        \includegraphics[width=\textwidth]{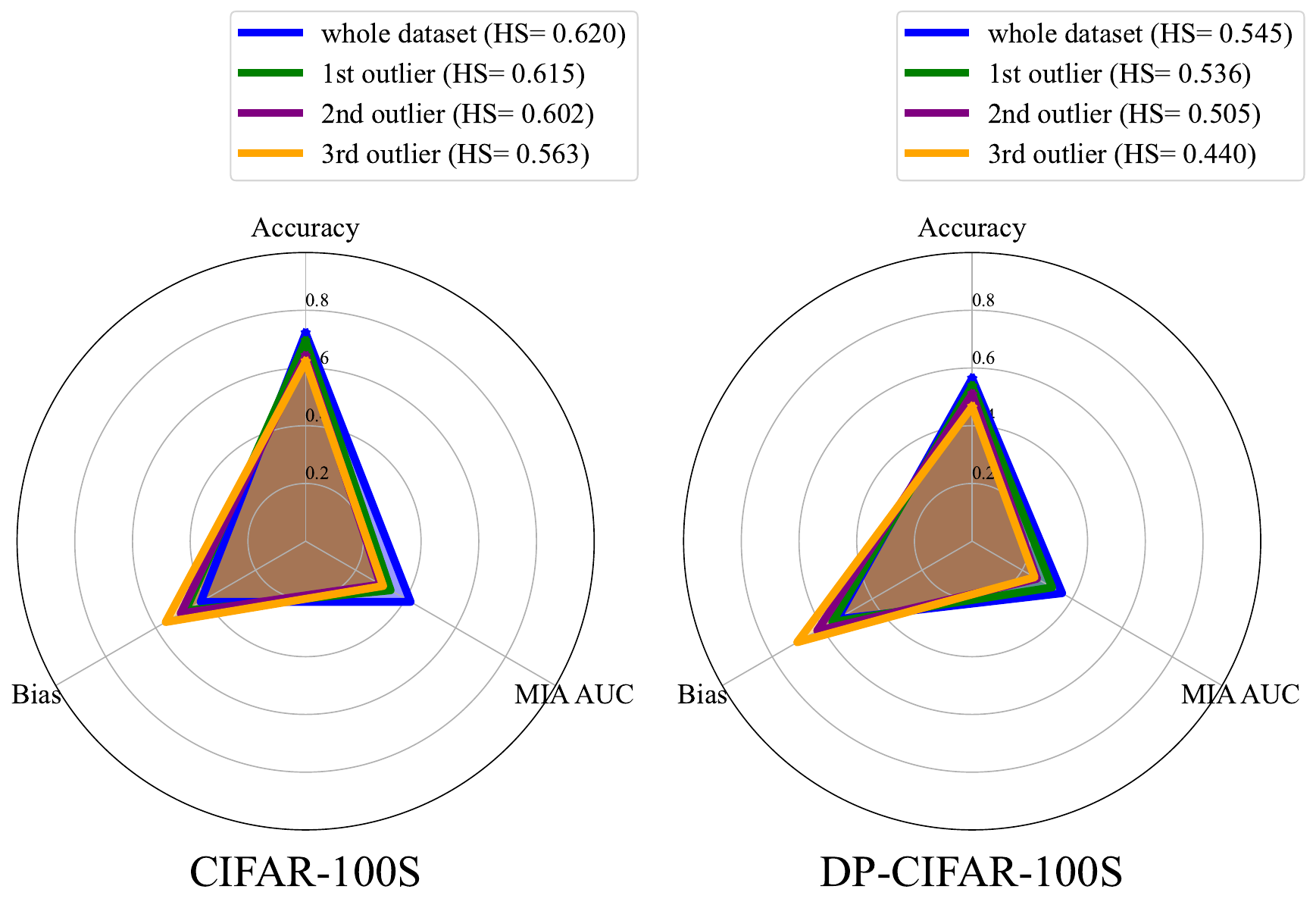}
        \caption{}
        \label{fig:abe_onion_effect_cifar100}
    \end{subfigure}
    \caption{The impact of removing outliers on model performance (accuracy, MIA AUC, bias, and HS) trained with and without DP on CIFAR-100 and CIFAR100S. DP-CIFAR-100 and DP-CIFAR-100S are used to denote when DP is applied.}
    \label{fig:onion_effectcifar100}
\end{figure*}

Figure~\ref{fig:onion_effectcifar100} evaluates the effect of outlier layers elimination on the performance metrics (accuracy, MIA AUC, bias, and HS) of models trained with $(8, 10^{-5})$-DP and without
DP on CIFAR-100 and CIFAR-100S. Figure~\ref{fig:accuracy_onion_effect_cifar100} indicates that removing outlier layers reduces accuracy by depriving the model of information needed for more accurate prediction. Figure~\ref{fig:attack_acc_onion_effect_cifar100} shows MIA AUC remains stable or slightly decreases across models, demonstrating the Onion Effect. Figure~\ref{fig:bias_onion_effect_cifar100} suggests outlier removal might amplify bias, as it decreases accuracy, particularly for underrepresented groups. Figure~\ref{fig:abe_onion_effect_cifar100} illustrates that removing outliers fails to balance accuracy, privacy, and bias in both private and non-private learning models.
\subsection{Ablation Study on Generalization Techniques}
Table~\ref{cifar-privacy-gt} summarizes the effect of generalization techniques individually on accuracy, MIA AUC, GGap, and bias for CIFAR-10 and CIFAR-10S datasets in private and non-private settings. The average accuracy for the COLOR and GRAY datasets has been reported for both CIFAR-10S and DP-CIFAR-10S. The key observations from the table indicate that generalization techniques like OBS and AM significantly improve model accuracy and reduce GGap across both DP and non-DP settings. DP settings tend to lower accuracy and also decrease MIA AUC. Additionally, almost all generalization techniques lead to a slight increase in bias.
\begin{table*}
\centering
\caption{An ablation study on the effect of generalization techniques on accuracy (acc), MIA AUC (AUC), GGap, and bias (if applicable) on CIFAR-10 and CIFAR-10S. DP-CIFAR-10 and DP-CIFAR-10S refer to these datasets used in the private setting under $(8, 10^{-5})$-DP.} 

\label{cifar-privacy-gt}
\begin{tabular}{|l|c|c|c|c|c|c|c|c|c|c|c|c|c|c|} 
\hline
\multirow{2}{*}{} & \multicolumn{3}{c|}{DP-CIFAR-10}                           & \multicolumn{3}{c|}{CIFAR-10}                              & \multicolumn{4}{c|}{~DP-CIFAR-10S}                         & \multicolumn{4}{c|}{CIFAR-10S}                               \\ 
\cline{2-15}
                  & acc                                             & AUC&GGap & acc                                             & AUC&GGap & acc                               & AUC &GGap&bias& acc                            & AUC&GGap & bias \\ 
\hline
Baseline~         & 49.47  & 0.55  &  3.76 & 71.13  & 0.55 & 1.39 & 53.74   & 0.85  & 8.93 &0.04& 66.25   & 0.85  &   4.13 &0.01\\ 
\hline
Baseline + GN & 54.96  & 0.55   & 2.48 & 76.08  & 0.55  & 1.08  & 56.28   & 0.84 &  7.26& 0.06 & 66.70  & 0.85  &  3.86 & 0.03\\ 
\hline
Baseline + OBS & 68.92  & 0.56  & 1.93  & 78.47   & 0.57 &  0.87  & 58.54  & 0.84  &  6.83&0.07 & 68.39   & 0.84   & 3.29 & 0.03\\ 
\hline
Baseline + WS & 52.42  & 0.54   & 2.26 & 73.19   & 0.58  & 1.17  & 55.62   & 0.83  & 7.78& 0.05 & 67.19    & 0.84  &  3.56 & 0.03\\ 
\hline
Baseline + AM & 54.38 & 0.55  & 2.41  & 79.83  & 0.55  &  0.54 & 55.27   & 0.82  &  7.62&0.05 & 70.87   & 0.85 &  2.75 & 0.04\\ 
\hline
Baseline + PA & 50.97  & 0.54  & 3.18  & 72.26  & 0.55  & 1.24  & 57.42  & 0.83 &  7.65& 0.05 & 73.84  & 0.86   &  1.89& 0.05\\ 
\hline
Baseline + SAT & 51.29 & 0.54  &  2.28 & 72.84 & 0.55 &  1.13 & 57.34  & 0.84  & 7.38& 0.06 & 69.04 & 0.82  &  2.16 & 0.04 \\
\hline
\end{tabular}
\end{table*}

\end{document}